\documentclass[11pt]{article}

\usepackage[preprint]{acl}

\usepackage{times}
\usepackage{latexsym}

\usepackage[T1]{fontenc}

\usepackage[utf8]{inputenc}

\usepackage{microtype}

\usepackage{inconsolata}

\usepackage{graphicx}

\usepackage{amsmath}
\usepackage{float}
\usepackage{subcaption}

\usepackage{orcidlink}
\usepackage{booktabs}

\usepackage{algorithm}
\usepackage{algpseudocode}
\usepackage{tabularx}
\usepackage{hyperref}

\title{Recursive Offline Distillation via Entropy Gain Sampling and Answer Mixing}

\author{
  \textbf{Andrey Goncharov\textsuperscript{1}\orcidlink{0009-0006-9471-2181}},
  \textbf{Daniil Vyazhev\textsuperscript{1}\orcidlink{0009-0004-4521-9062}},
  \textbf{Simon Epanov\textsuperscript{1}},
  \textbf{Alexey Zaytsev\textsuperscript{1}\orcidlink{0000-0002-1653-0204}},
  \\
  \\
  \textsuperscript{1}Applied AI Institute
  \\
  \small{
    \textbf{Correspondence:} \href{mailto:andrey@faillearnrepeat.net}{andrey@faillearnrepeat.net}
  }
}

\begin{document}
\maketitle
\begin{abstract}
  Distillation of reasoning chains from Large Language Models (LLMs) into their smaller siblings is a standard training approach.
  Frequently, researchers do not have full access to a teacher model and its internal representations, which limits the resulting performance.
  We propose a recursive offline distillation method combining two components: entropy-based sampling for selective distillation and answer mixing.
  Specifically, for three open models (3-4B), at each epoch, we sample reasoning chains weighted by entropy difference with an auxiliary medium model (70-72B) and, next, supplement thinking traces with single-token ground truth answers.
  Our approach consistently outperforms naive distillation with comparable data budgets ($57.15 \pm 1.01$ vs 51.61 on average) and offers a much steeper learning curve.
  Additionally, we present a comparative analysis of methods to generate and normalize distillation traces.
  We publish our code and data\footnote{\url{https://github.com/LabARSS/recursive_offline_distillation}}.
\end{abstract}

\section{Introduction}

Modern LLMs come in different sizes, and their performance is often correlated with the number of parameters.
The largest of them demonstrate exceptional results and span 1-2 trillion parameters.
Further domain adaptation \citep{ICLR2024_d51cd79a,lu2024finetuninglargelanguagemodels} or even inference at this scale requires massive computational resources \citep{wan2024efficientlargelanguagemodels} unavailable to most practitioners.
It prompts the need for capable smaller models \citep{chen2026rolesmallmodelsllm} that can be fine-tuned and deployed in resource-constrained environments.

\begin{figure}
  \centering
  \includegraphics[width=1\linewidth]{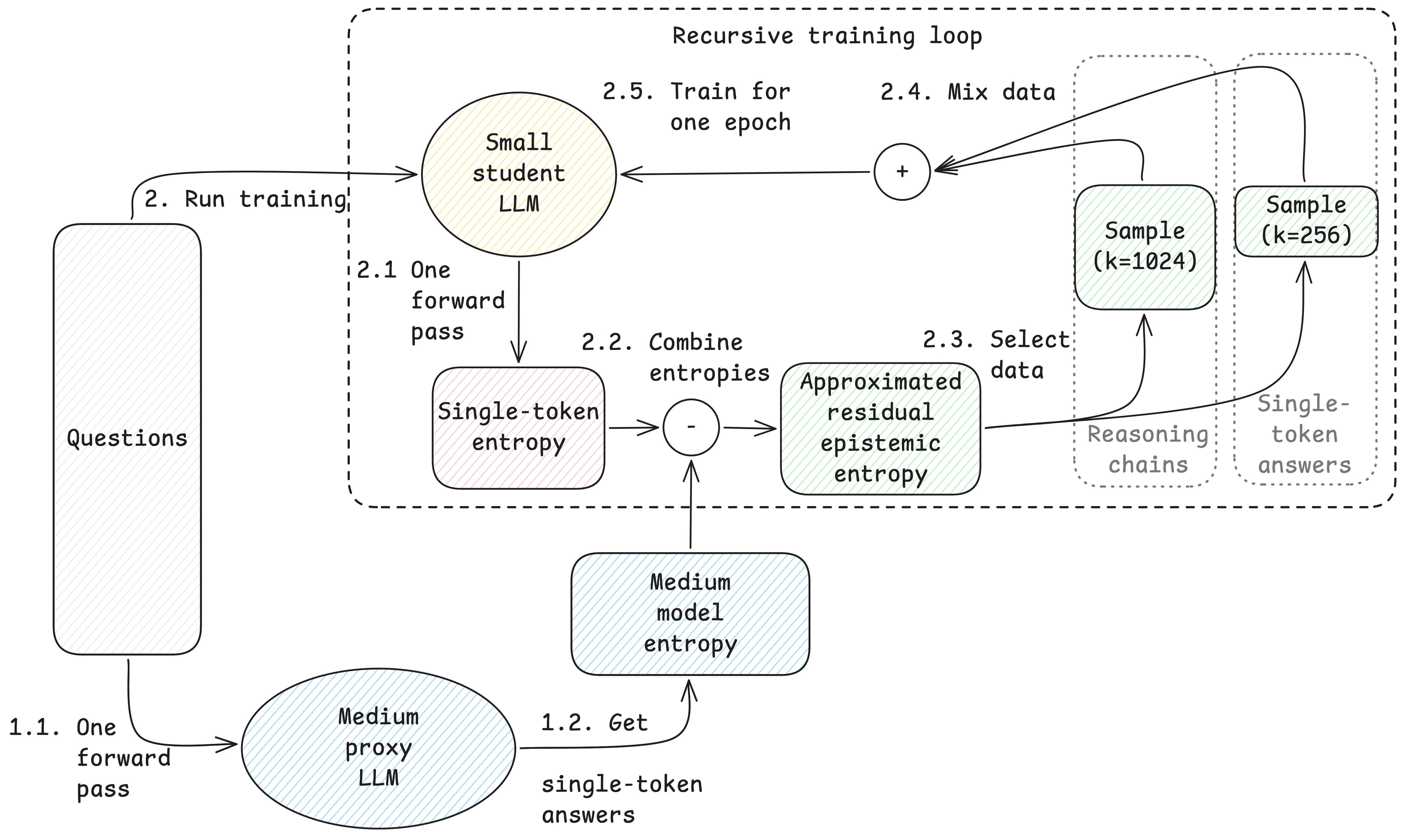}
  \caption{Recursive offline distillation scheme}
  \label{fig:recursive_caft_scheme}
\end{figure}

A prominent approach for small model training is distillation from a larger teacher model \citep{hinton2015distillingknowledgeneuralnetwork, sanh2020distilbertdistilledversionbert, xu2024surveyknowledgedistillationlarge}.
Many methods rely on having access to the online teacher \citep{ICLR2024_5be69a58, jiao-etal-2020-tinybert, ICLR2024_8ac015d4}, which is not always possible \citep{zhou-etal-2023-bridging, ryskulov2026efficientknowledgedistillationllms}.
It forces practitioners to explore offline distillation, where the student model is trained on the teacher's pre-generated text output \citep{li-etal-2023-symbolic, hsieh-etal-2023-distilling}.

We propose a fully automated recursive pipeline for offline distillation (Figure \ref{fig:recursive_caft_scheme}, described in \ref{sec:method}) based on entropy sampling and answer mixing.
The method consistently outperforms naive distillation ($57.15 \pm 1.01$ vs 51.61 on average), and it is designed to work in resource-constrained environments.

We contribute a family of entropy-based samplers and a lightweight data mix. Additionally, we provide a comparative analysis of augmentation strategies for synthetic reasoning chains when the teacher model does not perform perfectly in the target distillation domain and conduct sensitivity studies of the synthetic trace generation and normalization methods and sampling strategies.

\section{Related Work}

Recent work improved distillation through training objectives \citep{pmlr-v235-ko24c,xu-etal-2026-harnessing}, data selection \citep{pmlr-v306-just26a,li2025naturalthoughtsselectingdistillingreasoning,yang-etal-2026-select2reason}, and curriculum learning \citep{jiang-etal-2025-teach,zhang-liu-2026-calibrated}.
Joint supervision with rationales and labels improves data efficiency \citep{hsieh-etal-2023-distilling}.
Curriculum benefits vary with model capability, task complexity, and the choice of difficulty metric \citep{jia-etal-2026-makes}.
ELAD \citep{zhang-etal-2024-elad} selects uncertain student explanations for teacher revision.
Complexity-aware fine-tuning \citep{goncharov2026complexityawarefinetuning} selects reasoning examples by single-token entropy, but only once.
We recursively sample pre-generated traces using student-proxy entropy differences.

Trace synthesis includes explaining known solutions \citep{li2024distillingalgorithmicreasoning,wang2025reverseengineeredreasoningopenendedgeneration} and learning from corrections \citep{an2024learningmistakesmakesllm, ICLR2025_0967d7c8}.
We compare direct generation and various augmentations on the same task under constrained trace lengths.

\section{Method}
\label{sec:method}

\paragraph{Entropy sampling.} \citet{goncharov2026complexityawarefinetuning} show that a single-token entropy (Figure \ref{fig:single_token_entropy_illustration}) is a reasonable complexity predictor for a given entry in a question-answering dataset.
We assume that it approximates the total uncertainty.
With scaling laws \citep{kaplan2020scalinglawsneurallanguage} in mind, we hypothesize that larger models should have lower total uncertainty.
Therefore, in comparison to a small student model (3-4B), its larger sibling (proxy) (70-72B\footnote{This particular size is chosen based on the imposed resource constraints (80GB VRAM) and open model availability}) could have higher confidence (lower entropy) for a variety of data entries, and help the student to identify a useful knowledge gap.
We propose \emph{entropy gain} $H_S(x)-H_P(x)$ (similar to \citet{pmlr-v162-mindermann22a, lin2025rho1tokensneed}) as the primary heuristic, but evaluate alternatives as well.

The heuristic is inspired by epistemic-aleatoric uncertainty decomposition \citep{wang2025aleatoricepistemicexploringuncertainty, pmlr-v80-depeweg18a, malinin2021uncertaintyestimationautoregressivestructured}.
For a question $x$, assume $H(x)=A(x)+E(x)$, where $H_S$, $H_P$ - single-token entropies (student and proxy), $A$ - shared aleatoric component, and $E_S$, $E_P$ - model-specific epistemic components.
If its residual epistemic uncertainty satisfies $|E_S(x) - E_P(x)|\leq\varepsilon(x)$, then $|H_S(x)-H_P(x)|\leq\varepsilon(x)$.
Further discussion in Appendix \ref{sec:uncertainty_proxy_diagnostics}.

\paragraph{Answer Mixing.} We supplement 1024 reasoning chains with 256 single-token answer examples per training epoch.
The mix preserves model's capabilities for non-thinking answers, which is central to entropy sampling, and offers a surprising substantial performance uplift.

\begin{algorithm}
  \caption{Recursive offline distillation via entropy gain sampling and answer mixing}
  \label{alg:recursive_distillation}
  \small
  \begin{algorithmic}[1]
    \Require Student $S$, proxy model $P$, epochs $T$
    \Require Questions $D$, ground-truth answers $y_x$, and precomputed reasoning traces $r_x$
    \Ensure Trained student $S$
    \ForAll{$x\in D$}
    \State $H_P(x)\gets\Call{Entropy}{P,x}$
    \EndFor
    \For{$t=1,\ldots,T$}
    \ForAll{$x\in D$}
    \State $H_S(x)\gets\Call{Entropy}{S,x}$
    \State $\mathrm{EG}(x)\gets\max\{H_S(x)-H_P(x),0\}$
    \EndFor
    \State $Q\gets\Call{WeightedSample}{D,\mathrm{EG},1024}$
    \State $Q_A\gets\text{first }256\text{ questions in }Q$
    \State $B_R\gets[(x,r_x):x\in Q]$
    \State $B_A\gets[(x,y_x):x\in Q_A]$
    \State $S\gets\Call{Train}{S,B_R\mathbin{\Vert}B_A}$
    \EndFor
    \State \Return $S$
  \end{algorithmic}
\end{algorithm}

\paragraph{The proposed pipeline} consists of the following steps (Figure \ref{fig:recursive_caft_scheme}, Algorithm \ref{alg:recursive_distillation}):
(1) prompt the larger model (proxy) to answer all questions in the dataset with a single token once (or an ensemble, averaging entropies), collect entropy;
(2) for each training epoch,
(2.1) prompt the student model to answer all questions in the dataset with a single token, collect entropy;
(2.2) calculate entropy gain $EG(x) = \max\{H_S(x)-H_P(x),0\}$;
(2.3.1) sample 1024 synthetic reasoning chains without replacement, weighted by positive entropy gain;
(2.3.2) sample 256 single-token answers similarly, as a subset of the previous sample;
(2.4) concatenate samples;
(2.5) train the student using LoRA \citep{hu2021loralowrankadaptationlarge} on reasoning traces first, followed by single-token responses.

\section{Results}

\begin{figure*}[t]
  \centering
  \begin{subfigure}[b]{0.34\linewidth}
    \centering
    \includegraphics[width=1\linewidth]{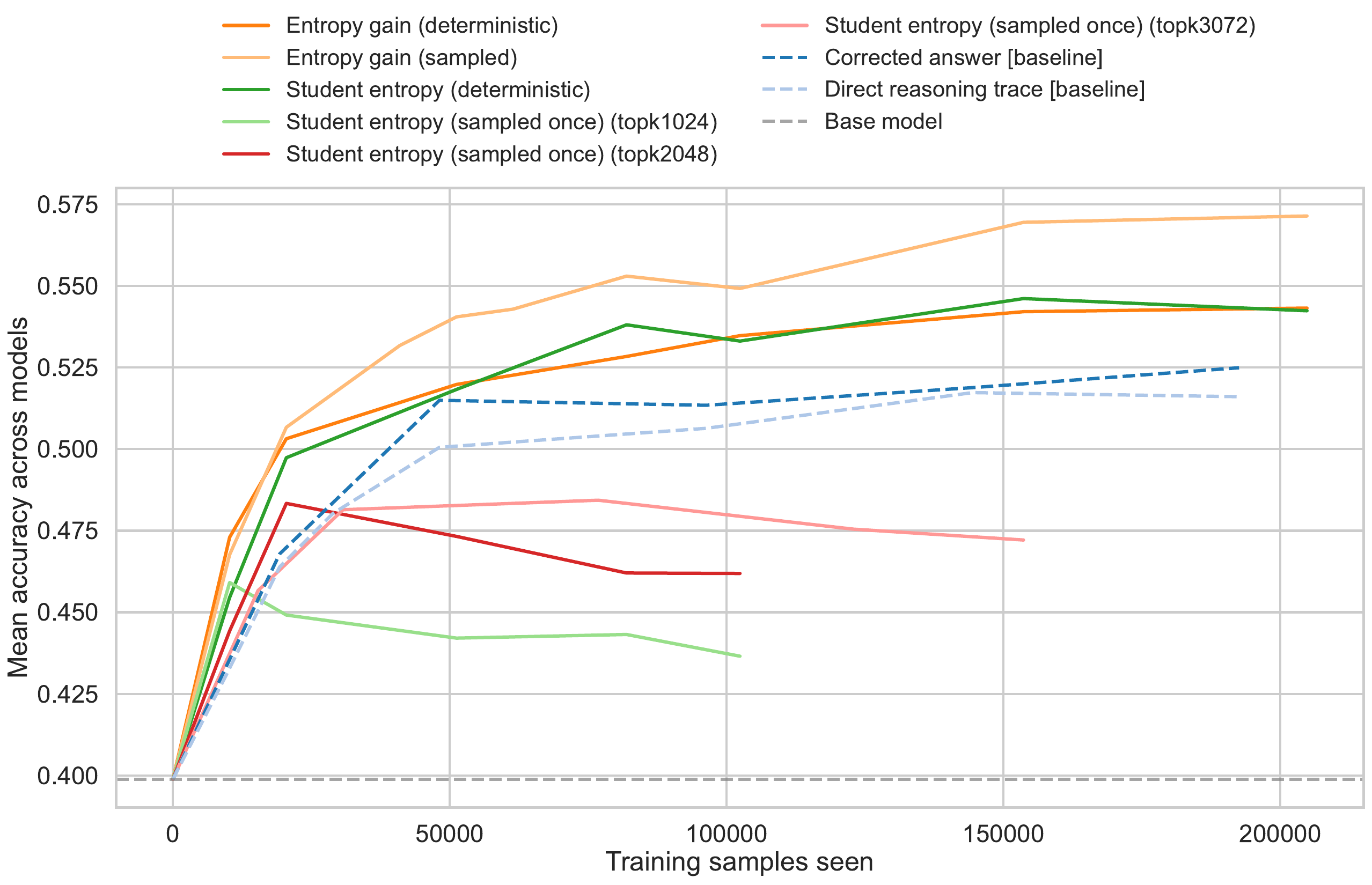}
    \caption{4096-token reasoning cap}
  \end{subfigure}
  \hfill
  \begin{subfigure}[b]{0.34\linewidth}
    \centering
    \includegraphics[width=1\linewidth]{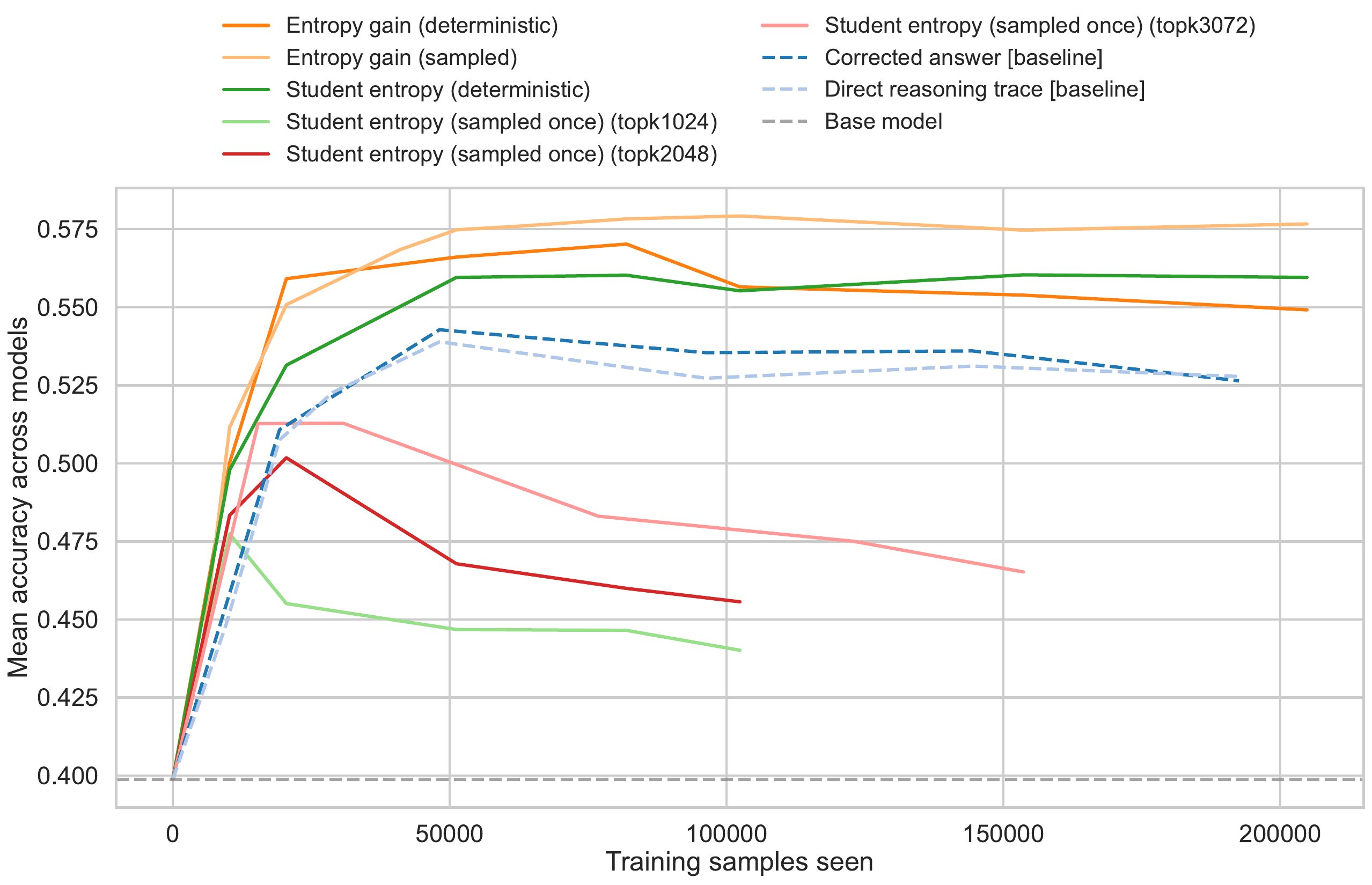}
    \caption{2048-token reasoning cap}
  \end{subfigure}
  \begin{subfigure}[b]{0.29\linewidth}
    \centering
    \includegraphics[width=1\linewidth]{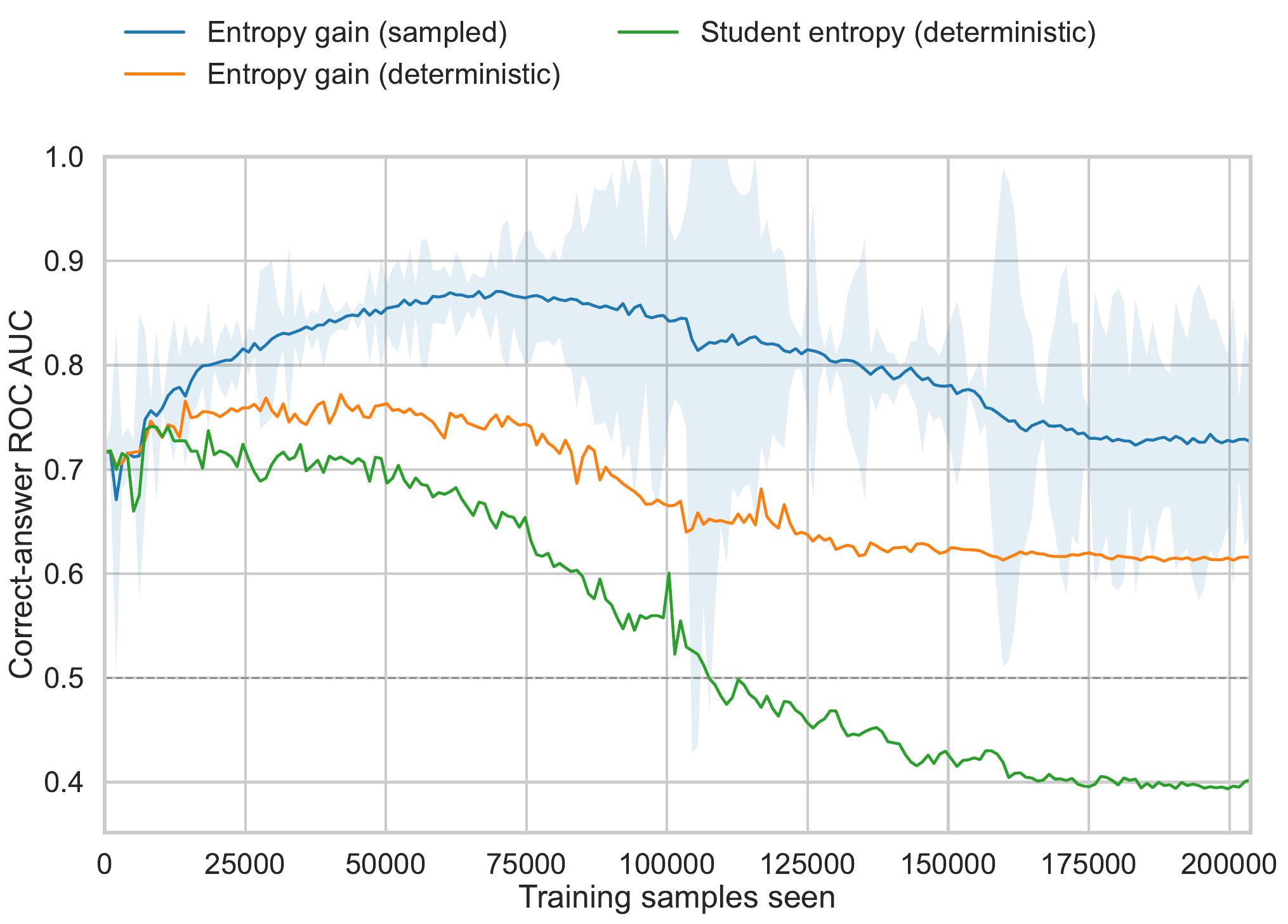}
    \caption{Student entropy ROC AUC}
  \end{subfigure}
  \caption{Distillation pipeline results across three models (average) on MMLU-Pro}
  \label{fig:main_results_ghraphs}
\end{figure*}

\begin{table*}[t]
  \centering
  \small
  \setlength{\tabcolsep}{4pt}
  \begin{tabular}{l *{5}{l}}
    \hline
    Method & Answer Mix & \multicolumn{1}{c}{LLaMA 3.2 3B} & \multicolumn{1}{c}{Qwen 2.5 3B} & \multicolumn{1}{c}{Phi4-mini} & \multicolumn{1}{c}{Mean} \\
    \hline
    Direct reasoning trace & No & 46.09 / 45.55 & 51.45 / 50.37 & 57.27 / 62.43 & 51.61 / 52.78 \\
    Corrected answer & No & 47.80 / 47.26 & 50.87 / 49.42 & 58.81 / 61.26 & 52.49 / 52.65 \\
    Student entropy (deterministic) & Yes & 48.84 / \underline{49.09} & \underline{55.86} / \underline{55.49} & 58.02 / \underline{63.30} & 54.24 / \underline{55.96} \\
    Entropy gain (deterministic) & Yes & \underline{49.58} / 49.00 & 53.16 / 53.33 & \underline{60.22} / 62.43 & \underline{54.32} / 54.92 \\
    Student entropy (once, $k=1024$) & No & 41.35 / 40.11 & 39.44 / 39.07 & 50.17 / 52.87 & 43.65 / 44.01 \\
    Student entropy (once, $k=2048$) & No & 42.19 / 41.06 & 43.18 / 43.02 & 53.20 / 52.62 & 46.19 / 45.57 \\
    Student entropy (once, $k=3072$) & No & 42.02 / 42.81 & 42.85 / 40.86 & 56.77 / 55.90 & 47.22 / 46.52 \\
    Entropy gain (sampled) & Yes &
    \begin{tabular}[c]{@{}l@{}}\textbf{52.58} $\pm\,1.94$ /\\\textbf{52.08} $\pm\,2.28$
    \end{tabular} &
    \begin{tabular}[c]{@{}l@{}}\textbf{56.64} $\pm\,3.79$ /\\\textbf{56.86} $\pm\,2.32$
    \end{tabular} &
    \begin{tabular}[c]{@{}l@{}}\textbf{62.22} $\pm\,2.42$ /\\\textbf{64.07} $\pm\,4.52$
    \end{tabular} &
    \begin{tabular}[c]{@{}l@{}}\textbf{57.15} $\pm\,1.01$ /\\\textbf{57.67} $\pm\,1.19$
    \end{tabular} \\
    \hline
  \end{tabular}
  \caption{Distillation pipeline last-epoch accuracy (4096-token / 2,048-token thinking budget) on MMLU-Pro.}
  \label{tab:distillation_main_results}
\end{table*}

\subsection{Experimental Setup}

\paragraph{Dataset:} Multiple-choice question-answering dataset MMLU Pro \citep{wang2024mmluprorobustchallengingmultitask}.
It has questions of different complexities from 14 domains: up to ten options each, one correct option. Random train/test split - 80/20.

\paragraph{LLMs:}
Proxy: ensemble of two medium models\footnote{All models and datasets are published under permissive licenses that allow them to be used for research purposes} (averaging their entropies) - LLaMA 3.3 70B \citep{grattafiori2024llama3herdmodels}, Qwen 2.5 72B \citep{qwen2.5}.
Students: Qwen2.5 3B \citep{qwen2.5}, Phi-4-Mini \citep{microsoft2025phi4minitechnicalreportcompact}, LLaMA 3.2 3B \citep{grattafiori2024llama3herdmodels}.
Reasoning trace source: DeepSeek V4 Pro and Flash \citep{deepseekai2026deepseekv4highlyefficientmilliontoken} via API.

\paragraph{Hardware:} We run all experiments\footnote{Training hyperparameters can be found in Appendix~\ref{sec:training_hyperparameters}.}\footnote{We used AI coding agents for a substantial part of the experimental and analytical code} on a single A100/H100\footnote{We had to switch cards depending on availability} GPU with 80GB of VRAM to validate the method in resource-constrained environments.

\subsubsection{Our method and results}

We evaluate the proposed pipeline for 200 epochs (205k questions seen) and 2 thinking budgets (2048 and 4096 tokens).
It is compared under similar data budgets with:\\
\emph{Direct reasoning trace, Corrected answer} - naive distillation on the most promising types of reasoning traces (see \ref{sec:synth_trace_analysis});\\
\emph{Student entropy (deterministic), Entropy gain (deterministic)} - alternative pipelines with weighted sampling replaced with greedy sampling;\\
\emph{Student entropy (once, k={x})} - distillation executed against k samples selected once by student entropy (not recursively at each epoch).

Notably, the proposed data mix increases the number of training tokens by less than 0.01\% (256 extra tokens compared to $\approx$2.5m) or 9\% accounting for the system prompt in the context ($\approx$250k extra tokens compared to $\approx$2,75m).

All randomly sampled methods use 3 random seeds, reporting $\pm$ 95\% confidence intervals, and deterministic methods are evaluated once with greedy sampling.

\subsection{Main Results}

Table \ref{tab:distillation_main_results} and Figure \ref{fig:main_results_ghraphs} show the results for MMLU Pro.
We see that the proposed method produces a significant improvement over naive distillation (up to $10\%$ gain).
For both reasoning budgets, it provides the best result on average across three models in comparison with other approaches.
Greedy sampling demonstrates subpar and less stable curves, though surpassing the baselines.
Greedy distillation based on student entropy sampled once \citep{goncharov2026complexityawarefinetuning} shows signs of severe degradation after the initial uplift revealing the fragility of the approach.

We also note that our method helps with the model's single-token entropy correctness discrimination, increasing ROC AUC for the first half of training.
Shortly after, it starts declining, which is loosely correlated with the method's plateau.

\begin{figure*}[t]
  \centering
  \begin{subfigure}[b]{0.31\linewidth}
    \centering
    \includegraphics[width=1\linewidth]{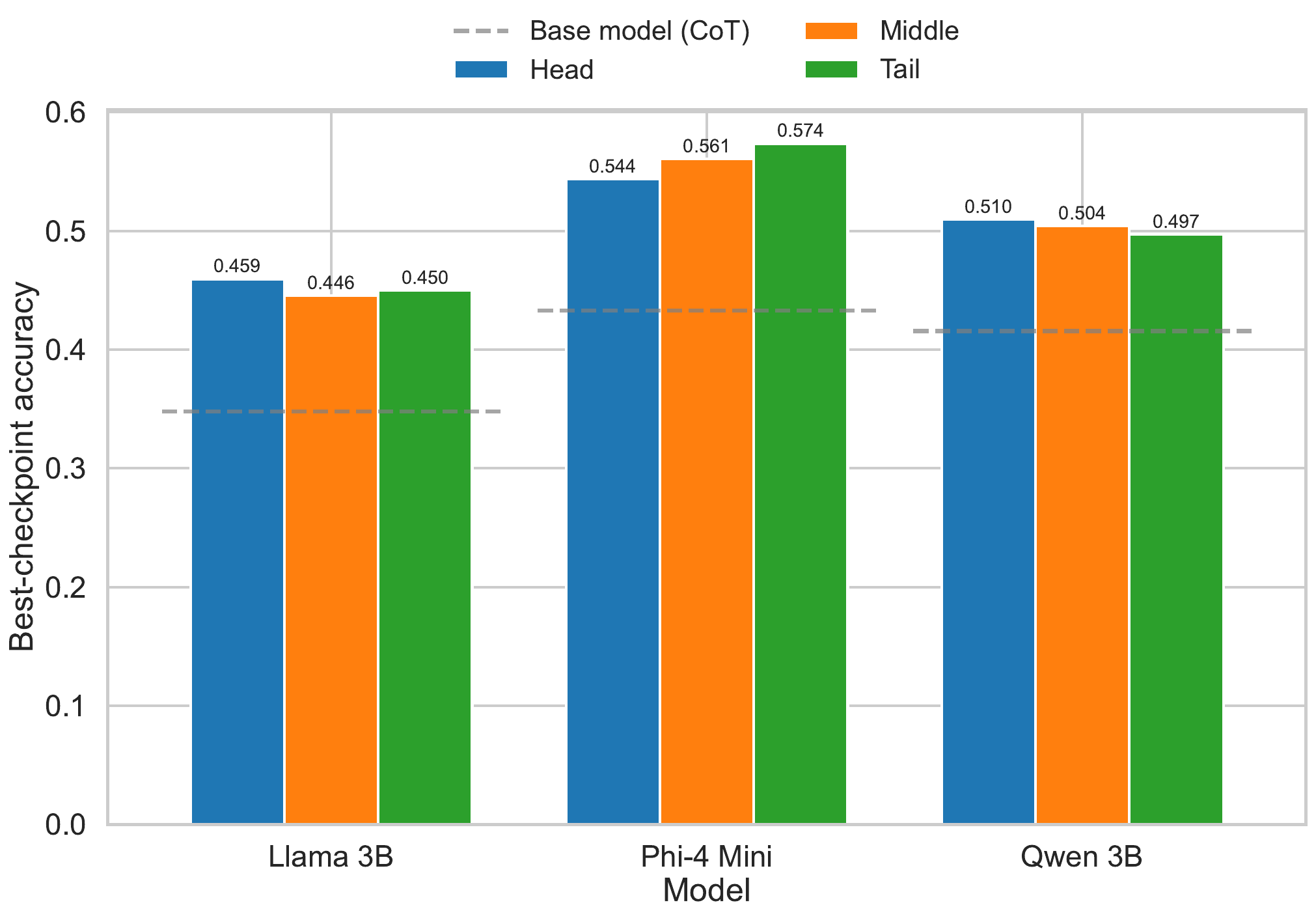}
    \caption{by truncation type}
  \end{subfigure}
  \hfill
  \begin{subfigure}[b]{0.35\linewidth}
    \centering
    \includegraphics[width=1\linewidth]{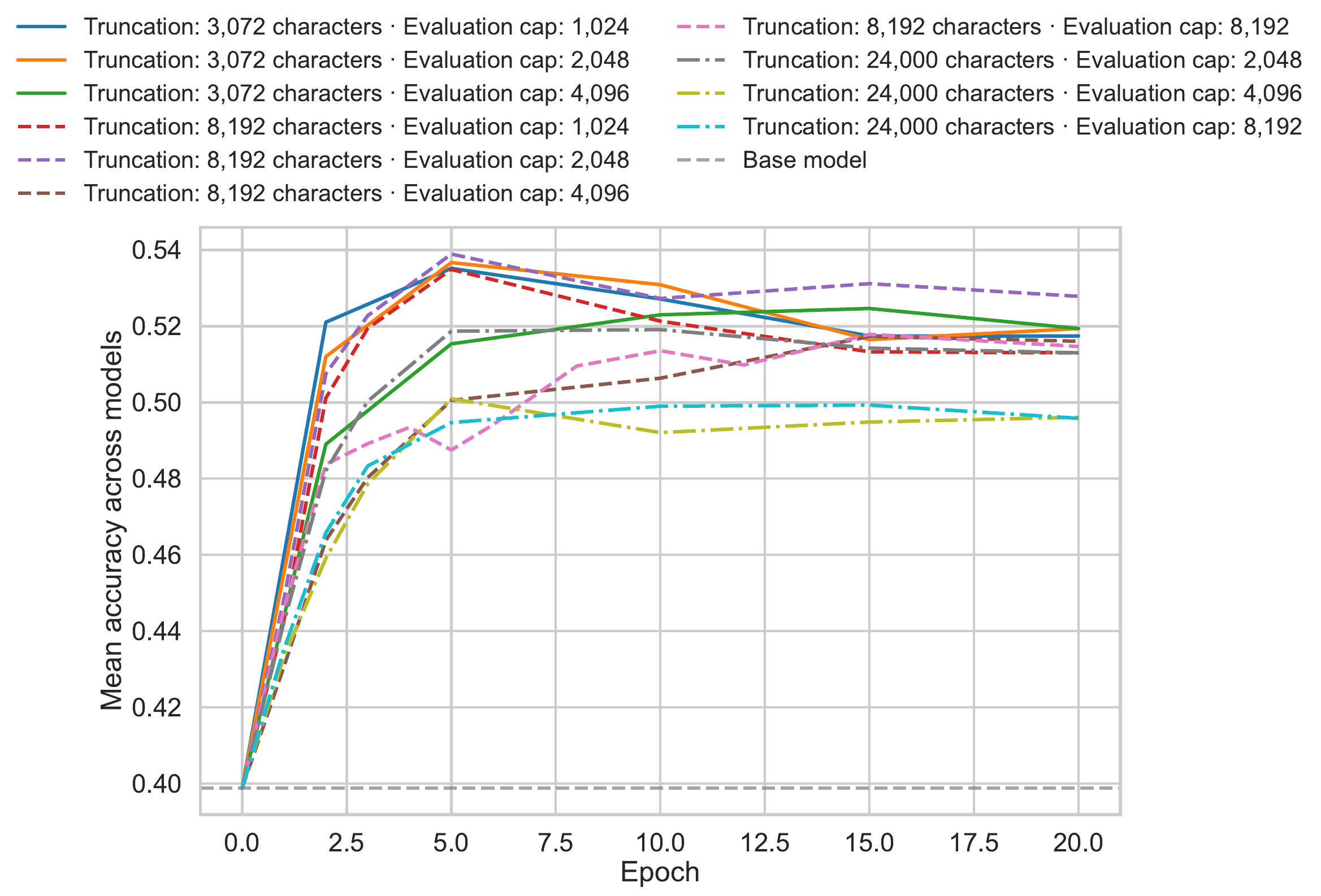}
    \caption{by length}
  \end{subfigure}
  \hfill
  \begin{subfigure}[b]{0.30\linewidth}
    \centering
    \includegraphics[width=1\linewidth]{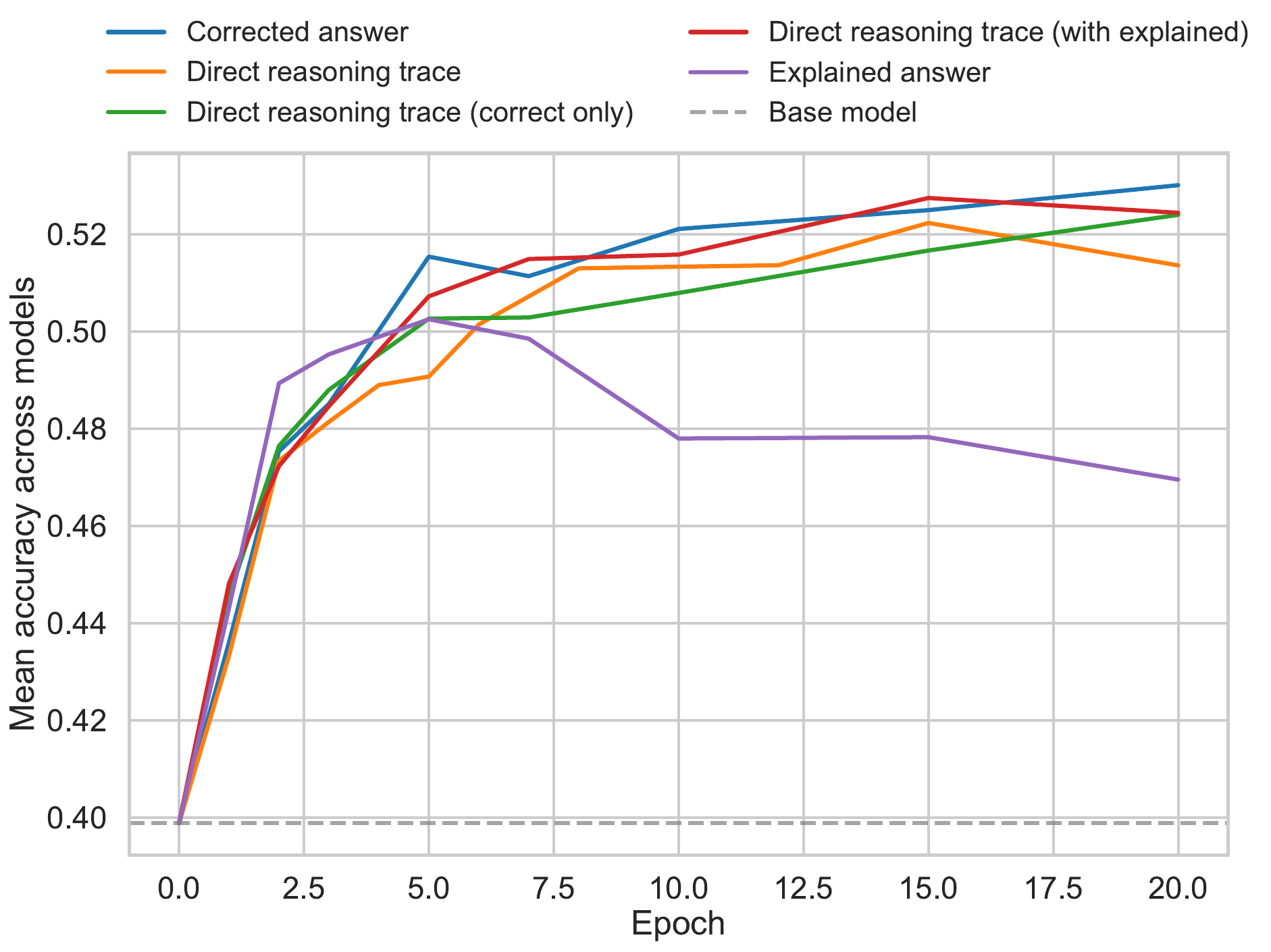}
    \caption{by generation method}
  \end{subfigure}
  \caption{Comparison of synthetic traces across three models (average)}
  \label{fig:synth_comparison}
\end{figure*}

\subsection{Synthetic Trace Generation}
\label{sec:synth_trace_analysis}

In resource-constrained environments, available VRAM imposes a strict limitation on the length of distilled reasoning chains.
Once we start truncating them, we face a new challenge: which part of the trace contains the most useful signal?
Finally, how exactly do we prompt the teacher model to generate reasoning traces?

We evaluate reasoning traces by:
(a) \emph{truncation type} - which part of the trace we keep when it exceeds the maximum length;
(b) \emph{length} - the optimal training truncation limit (characters) and evaluation thinking budget (tokens);
(c) \emph{generation method} (below).

Generation methods (details in Appendix~\ref{sec:synth_genration_methods_details}):
(1) \emph{Direct reasoning trace} - baseline prompt to just answer the question;
(2) \emph{Direct reasoning trace (correct only)} - same as (1), but we keep only traces with correct answers;
(3) \emph{Explained answer} - we give a correct answer to the model and prompt it to explain it;
(4) \emph{Direct reasoning trace (with explained)} - same as (1), but replace incorrect traces with (3);
(5) \emph{Corrected answer} - same as (1), but for incorrect traces, we give the model correct answers and ask it to update its reasoning.

Figure~\ref{fig:synth_comparison} shows that models have a slight preference for head truncation, medium to short training chains and thinking budgets.
Surprisingly, longer reasoning chains and greater thinking budgets lead to notable performance decreases.
As to the generation methods, we see that augmented reasoning traces lead to elevated performance, while the consistent data leakage results in degradation.
Based on the findings, we use 8,192-character head truncation and \emph{Corrected answer} generation for the pipeline.

\subsection{Sensitivity Study}

To attribute performance gains to either the proposed data mix or the entropy-based score and sampler, we compare \emph{entropy gain} with:
(1) alternative pipelines where we sample according to \emph{student entropy} and \emph{discounted entropy gain} (see below);
(2) an alternative pipeline where we use a random sampler, but with our data mix.

\emph{Discounted entropy gain} is defined as $\max\{H_S(x)- k(t) * H_P(x),0\}, k(t) \in [0, 1]$, where $t$ is the training epoch.
We set $k(t)$ initially to 0.8 and decrease it according to a schedule, hypothesizing that, as the student undergoes training, the static medium model's advantage disappears.

Table \ref{tab:sampler_ablation} shows that, across three models on average, entropy gain provides better results, which makes the heuristic promising, and we adopt it as the primary sampler for the method.
At the same time, random sampling performs exceptionally well (confirming findings of \citet{okanovic2023repeatedrandomsamplingminimizing}), which points to the proposed data mix as a major contributor to the gains.

\begin{table}
  \centering
  \small
  \setlength{\tabcolsep}{5pt}
  \begin{tabular}{l *{3}{l}}
    \hline
    Sampler & Mix & \multicolumn{2}{c}{Mean Accuracy} \\
    \hline
    Direct reasoning trace (baseline) & No & 51.61 & \\
    Corrected answer (baseline) & No & 52.49 & \\
    Random & Yes & 55.08 & $\pm\,0.80$ \\
    Student entropy & Yes & 55.87 & $\pm\,1.40$ \\
    Discounted entropy gain & Yes & \underline{56.06} & $\pm\,0.68$ \\
    Entropy gain & Yes & \textbf{57.15} & $\pm\,1.01$ \\
    \hline
  \end{tabular}
  \caption{Distillation pipeline last-epoch accuracy for different sampling methods (4096-token thinking budget)}
  \label{tab:sampler_ablation}
\end{table}

\section{Conclusion and Discussion}

This paper introduces a recursive offline distillation method rooted in entropy sampling and a lightweight answer mix ($57.15 \pm 1.01$ mean accuracy across three students on MMLU-Pro vs $51.61$ for direct distillation).
Experiments show that adding a small portion of direct answers to reasoning chains provides a major uplift. The entropy gain sampler contributes additional gains.
Corrected reasoning chains, and, surprisingly, \emph{shorter} traces and reasoning budgets perform better.
Exploration of more rigorous uncertainty estimation and nature of the data mix effect remains future work.

\section*{Limitations}

\begin{itemize}
  \item The method is tested only with small models. We do not know how well it is going to generalize to larger LLMs. Additionally, it might be challenging to find a suitable proxy for larger students.
  \item We use a single dataset. We do not know how well it translates to other settings.
  \item Alternative uncertainty estimations, such as ensemble agreement and others, are left unexplored.
  \item We did not test and compare multiple sizes of proxies and their impact on performance.
  \item Different compositions of the data mix are not tested due to lack of resources.
  \item We tested only a single schedule for \emph{Discounted Entropy Gain}. Alternatives remain to be explored.
  \item Some low-resource environments might have even more constraints and therefore struggle to fit a suitable proxy into VRAM.
\end{itemize}


\bibliography{custom}

\appendix

\label{sec:appendix}

\section{Experimental Hyperparameters}
\label{sec:training_hyperparameters}

\begin{center}
  \small
  \begin{tabularx}{\linewidth}{@{}p{0.48\linewidth}X@{}}
    \toprule
    Hyperparameter & Value \\
    \midrule
    Optimizer & AdamW (\texttt{adamw\_torch}) \\
    Learning rate & $10^{-4}$ \\
    Learning-rate schedule & Cosine \\
    Warmup ratio & 0.06 \\
    Weight decay & 0 \\
    Adam $(\beta_1,\beta_2)$ & $(0.9,0.999)$ \\
    Adam epsilon & $10^{-8}$ \\
    Maximum gradient norm & 1.0 \\
    Mixed precision & BF16 enabled \\
    Gradient checkpointing & Enabled \\
    \texttt{torch.compile} & Enabled \\
    LoRA rank $r$ & 8 \\
    LoRA alpha & 16 \\
    LoRA scaling $\alpha/r$ & 2 \\
    LoRA dropout & 0.1 \\
    Initialization & Gaussian \\
    \bottomrule
  \end{tabularx}
\end{center}

\section{Reporting format}
\label{sec:confidence_intervals}

For experiments with random sampling, for each model, we report the mean $\bar{x}$ across training seeds with a two-sided 95\% Student's $t$ confidence interval. For the mean accuracy across models, we weight model means equally and treat them as independent. 

For aggregate ROC AUC curves, we instead average all three models within each seed, then apply the Student's $t$ interval across these seed averages.

For deterministic experiments we report results from a single run.

\section{Reasoning distillation approach}
\label{sec:explicit_thinking_tokens}

Originally, all students are non-reasoning models. We adopt special tokens, \texttt{<think>} and \texttt{</think>}, to teach them separate reasoning from the final answers. We extend student's tokenizer and initialize corresponding embedding rows from the mean of the existing rows.

For reasoning evaluation, we append \texttt{<think>} to the assistant prompt. For single-token entropy estimation, we omit the thinking prefix and expect a model to answer with a single token without thinking. The 256 direct-answer examples in each mixed-training epoch help preserve this latter capability while the 1,024 reasoning examples teach reasoning mode.

\paragraph{Example.}
Consider the toy question 
\begin{quote}\small
  A rectangle has sides of length 3 and 2. What is its perimeter? a.\ 6; b.\ 10.
\end{quote}
A reasoning target is: 
\begin{quote}\small
  \texttt{<think>}There are two sides of each length, so the perimeter is $2(3+2)=10$.\texttt{</think>b}
\end{quote}
A single-token estimation target is:
\begin{quote}\small
  \texttt{b}
\end{quote}

\section{Accuracy gains beyond answer formatting}
\label{sec:mix_answer_formatting}

What do single-token answers contribute? Is it genuine quality uplift or merely formatting?

The evaluation budget only limits reasoning, rather than stopping the entire generation before the final answer. If the model reaches a reasoning cap, we forcefully insert \texttt{</think>} and allow model to finish generation of the final answer for extra ten tokens (to account for possible spaces, punctuation and for analysis).

We consider an answer as invalid if it is empty or is not among the options available for that question. A valid but incorrect option is an ordinary prediction error. The answer is normalized by removing special tokens,  stripping whitespace, case normalization.

Table~\ref{tab:answer_formatting_validity} compares the last checkpoints used in our main results: epoch 20 for direct distillation (no answer mixing) and epoch 200 for the entropy-sampling method (with answer-mixing).

\begin{table}[t]
  \centering
  \small
  \setlength{\tabcolsep}{3pt}
  \begin{tabular}{lr r rr}
    \toprule
    Student & Cap & No mix & Mix \\
    \midrule
    LLaMA 3.2 3B & 2,048 & 1 & 2 \\
    Qwen 2.5 3B & 2,048 & 1 & 0 \\
    Phi4-mini & 2,048 & 3 & 8 \\
    \midrule
    LLaMA 3.2 3B & 4,096 & 0 & 2 \\
    Qwen 2.5 3B & 4,096 & 3 & 0 \\
    Phi4-mini & 4,096 & 187 & 122 \\
    \bottomrule
  \end{tabular}
  \caption{Invalid extracted answers. No mix denotes direct reasoning-trace distillation; mix denotes entropy-gain sampling with direct-answer examples in the data mix.}
  \label{tab:answer_formatting_validity}
\end{table}

At the 2,048-token budget, the no-mix students already produce valid extracted answers nearly perfectly. Even correcting every invalid baseline answer would add an increment far below the accuracy gains in Table~\ref{tab:distillation_main_results}. At 4,096 tokens, we see similar picture for two models, and also observe that Phi4-mini has a comparable number of errors between the mix and non-mix runs. It points to answer-mixing genuinely improving model's quality, not merely formatting of the output.

\section{Single-token entropy}

\begin{figure}
  \centering
  \includegraphics[width=1\linewidth]{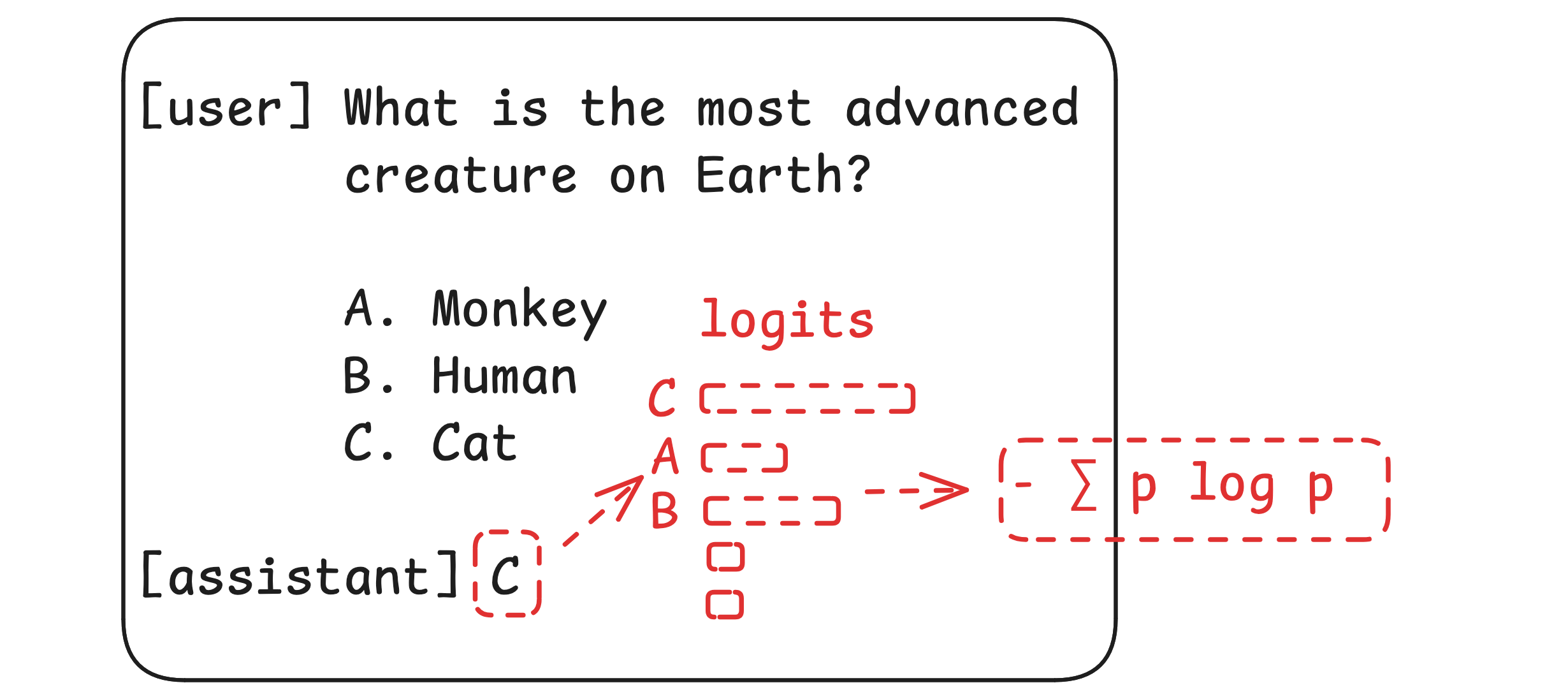}
  \caption{Single-token entropy example}
  \label{fig:single_token_entropy_illustration}
\end{figure}

Single-token entropy (Figure \ref{fig:single_token_entropy_illustration}) measures a model's uncertainty over the first answer token.
For a question $x$, we apply softmax to the first-token logits to obtain probabilities $p(v\mid x)$ over the model's full vocabulary $V$, and compute normalized Shannon entropy:
\begin{equation}
  H(x) = -\frac{1}{\log |V|}\sum_{v\in V} p(v\mid x)\log p(v\mid x).
\end{equation}
Dividing by $\log |V|$ puts student and proxy entropies on a common $[0,1]$ scale despite differences in vocabulary size.

\section{Synthetic trace generation methods}
\label{sec:synth_genration_methods_details}

\subsection{Direct reasoning trace.}
The teacher receives the question and options, generates reasoning, and predicts an option without seeing the ground-truth answer. The baseline retains both correct and incorrect responses.

\paragraph{System prompt}
\begin{quote}\small
  The following are multiple choice questions about \{subject\}. Choose a correct option letter. Answer with a single symbol. Do not print anything else.
\end{quote}

\subsection{Direct reasoning trace (correct only).}

We retain only baseline responses whose predicted option matches the ground truth. System prompt is as above.

\subsection{Explained answer.}
The teacher receives the question, options, and correct answer. It generates a step-by-step explanation without revealing it knows ground truth.

\paragraph{System prompt}
\begin{quote}\small
  The following is a multiple choice question about \{subject\}. The correct answer is provided to you, but your task is to produce a detailed chain-of-thought with step-by-step reasoning that reads as if you were solving the question from scratch---including any natural uncertainty, consideration of multiple options, or self-correction. Do not state or imply that the answer was given to you. End your response with Answer: \texttt{<letter>}.
\end{quote}

\subsection{Direct reasoning trace (with explained).}
We retain correct baseline responses and replace each incorrect response with the explained-answer response for the same question. 

\subsection{Corrected answer.}
For each incorrect baseline response, the teacher receives the original reasoning, its incorrect option, and the correct option. It continues the trace with the self-correction. Correct baseline responses remain intact.

\paragraph{System prompt}
\begin{quote}\small
  The following are multiple choice questions about \{subject\}. You will be shown a question, a partial chain-of-thought that was previously produced for it, the incorrect option that chain-of-thought arrived at, and the correct option letter. Continue the partial chain-of-thought from where it left off so that it naturally arrives at the correct option. Do not restart the reasoning, do not repeat what was already written, and do not acknowledge that the previous attempt was wrong or that the correct answer was given to you. Write the continuation as if you were the same reasoner noticing a mistake or new consideration mid-thought and revising course. In the end, answer with the correct option letter. End your response with Answer: \texttt{<letter>}.
\end{quote}

\subsection{User prompts}
All of them share the same user prompt:

\begin{quote}\small\ttfamily
  Question: \{question\}\\
  Options:\\
  a. \{option text\}\\
  b. \{option text\}\\
  \ldots
\end{quote}
For explained answers, append \texttt{Correct option:} followed by the ground-truth letter. For corrections, append:
\begin{quote}\small\ttfamily
  Partial reasoning so far:\\
  \{original reasoning\}\\
  Incorrect option this reasoning led to: \{predicted letter\}\\
  Correct option: \{ground-truth letter\}
\end{quote}

\subsection{Illustrative examples}
For brevity, consider a toy question with two options: ``A rectangle has sides of length 3 and 2. What is its perimeter? a.\ 6; b.\ 10.''

\paragraph{Direct trace.}
``The perimeter is $3\times2=6$. Answer: a.'' The baseline retains this response; the correct-only variant discards it.

\paragraph{Explained answer.}
Given correct option b, a fresh explanation is: ``There are two sides of each length, so the perimeter is $2(3+2)=10$. Answer: b.'' The with-explained variant substitutes this response for the incorrect direct trace.

\paragraph{Corrected answer.}
Given the original reasoning, incorrect option a, and correct option b, a continuation is: ``That product gives the area. For the perimeter, I need to add all four sides: $3+2+3+2=10$. Answer: b.'' The target retains the original reasoning, appends this continuation, and ends with b.

\subsection{Per-model results}
\label{sec:synth_per_model}
\begin{figure*}[t]
  \centering
  \begin{subfigure}[b]{0.32\linewidth}
    \centering
    \includegraphics[width=\linewidth]{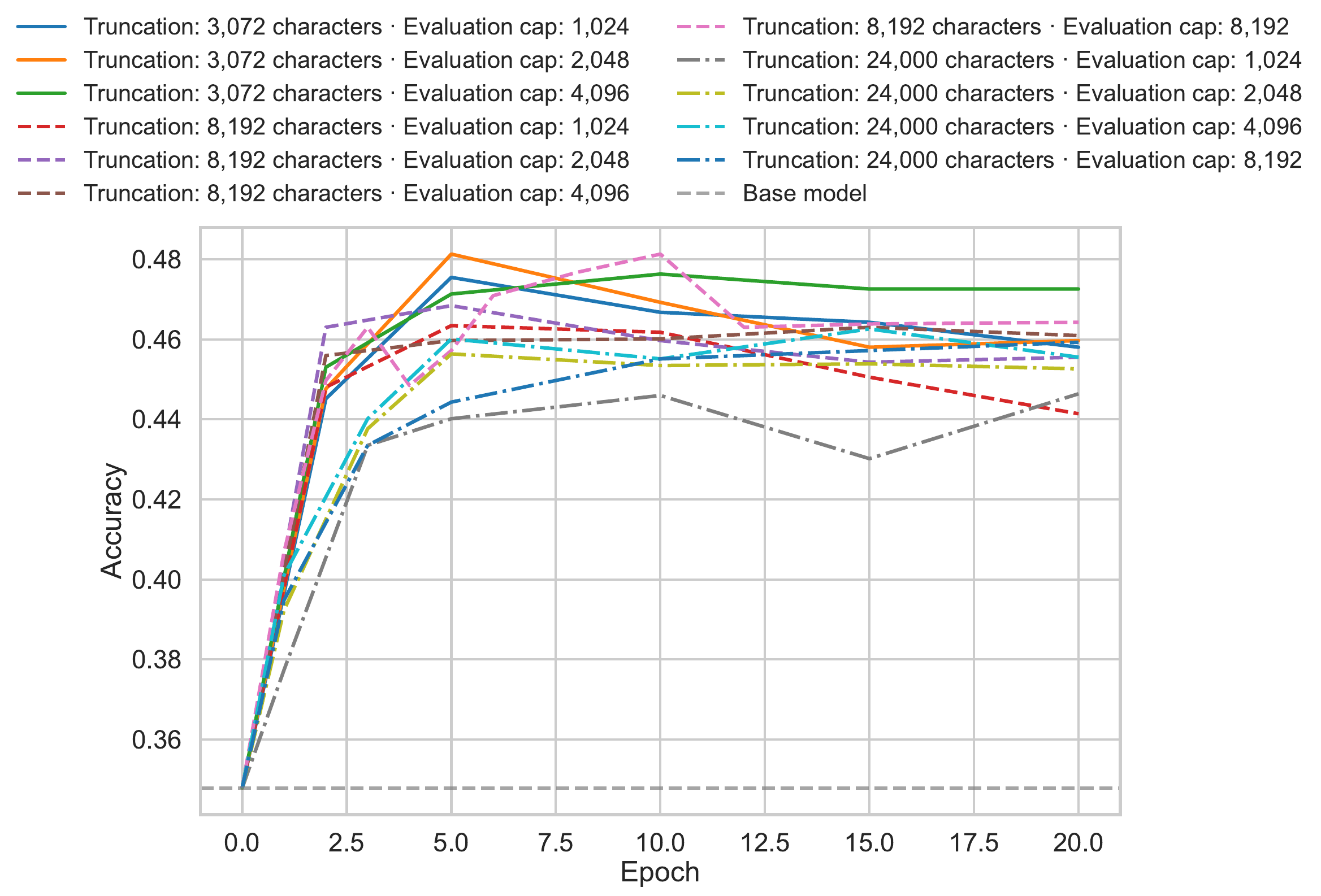}
    \caption{Head-truncation comparison}
  \end{subfigure}
  \hfill
  \begin{subfigure}[b]{0.32\linewidth}
    \centering
    \includegraphics[width=\linewidth]{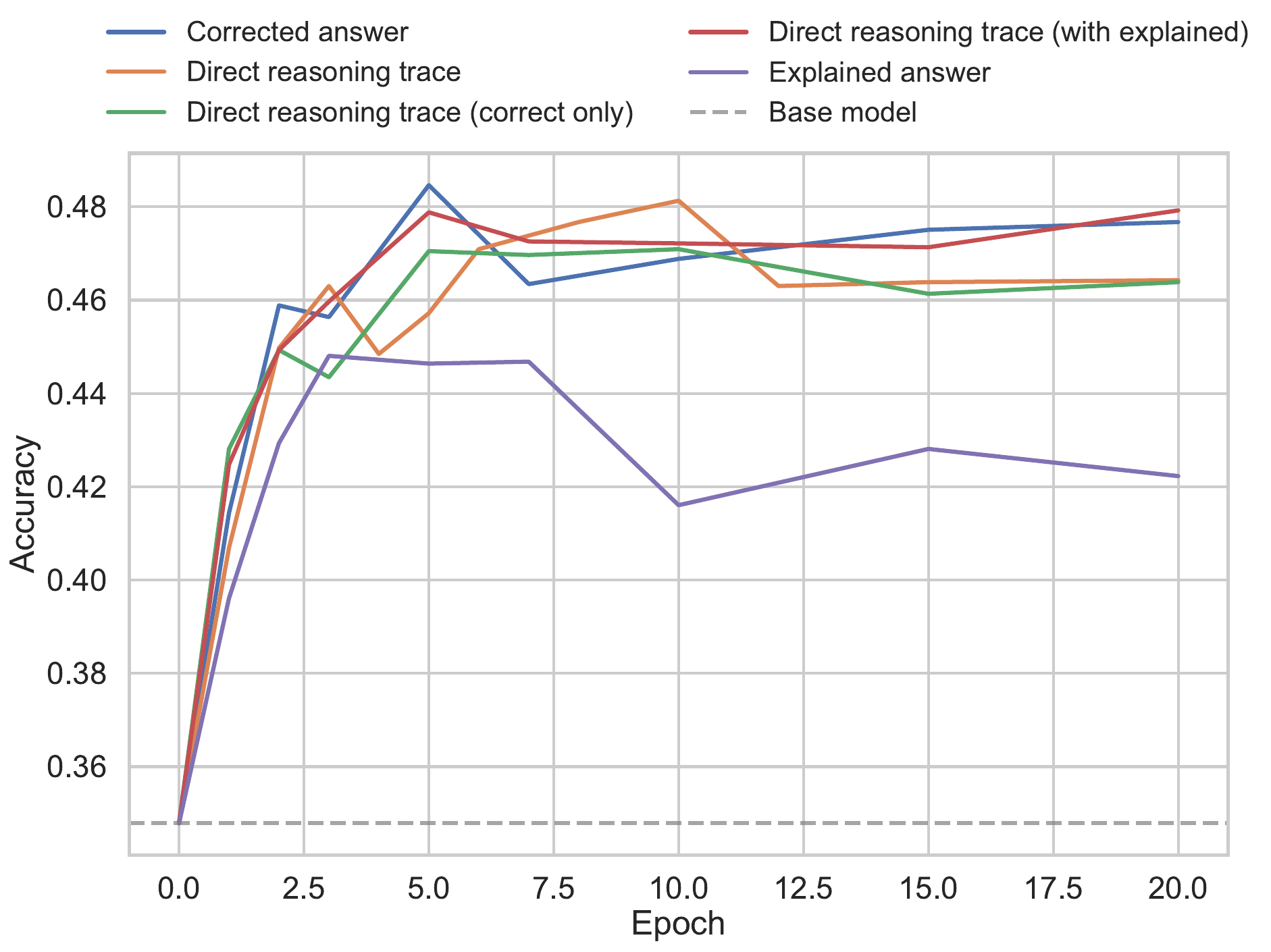}
    \caption{Generation methods: truncation 8,192, eval cap 8,192}
  \end{subfigure}
  \hfill
  \begin{subfigure}[b]{0.32\linewidth}
    \centering
    \includegraphics[width=\linewidth]{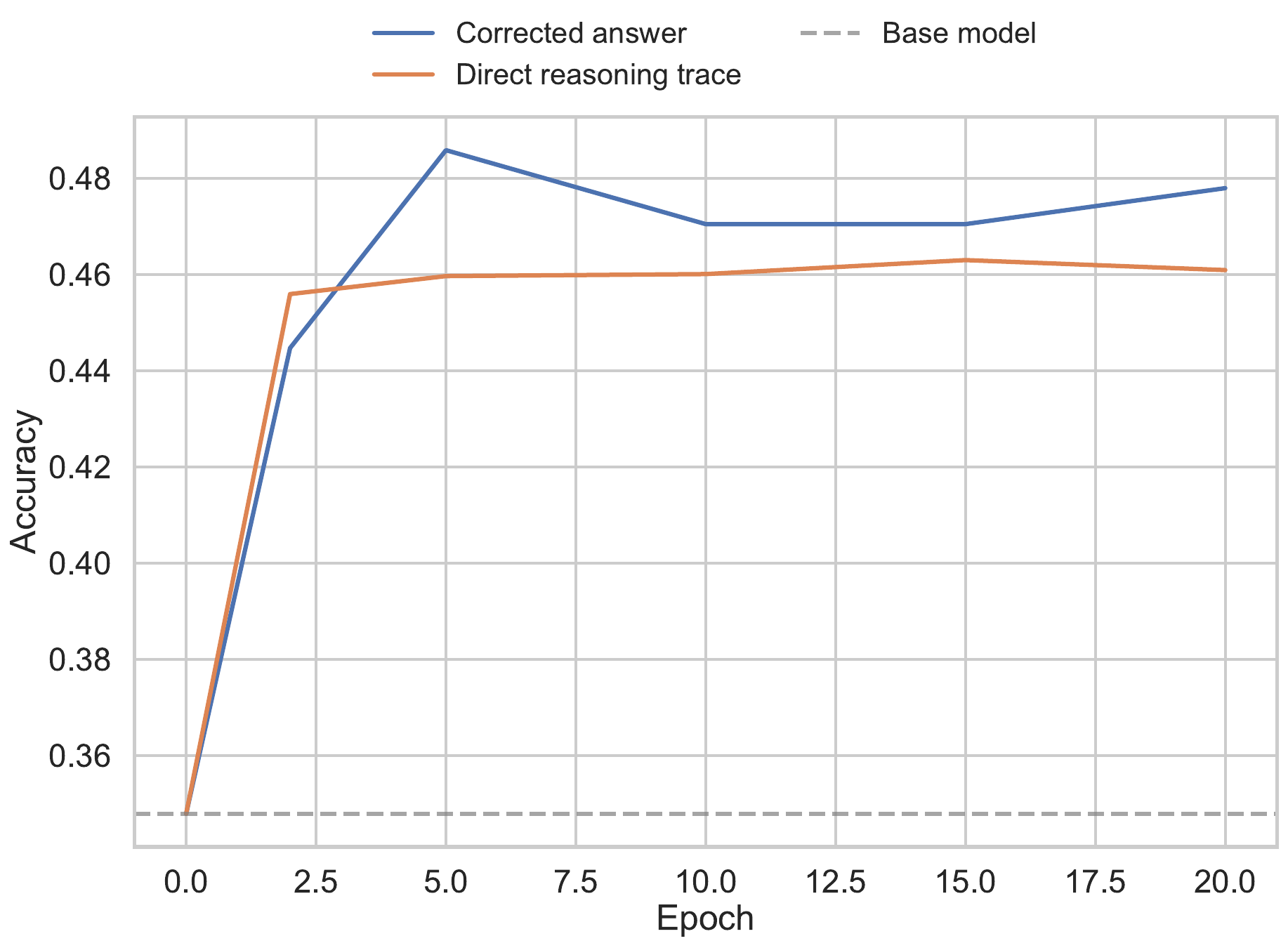}
    \caption{Generation methods: truncation 8,192, eval cap 4,096}
  \end{subfigure}
  \caption{Synthetic-trace distillation results for LLaMA 3.2 3B.}
  \label{fig:synth_llama_individual}
\end{figure*}

\begin{figure*}[t]
  \centering
  \begin{subfigure}[b]{0.32\linewidth}
    \centering
    \includegraphics[width=\linewidth]{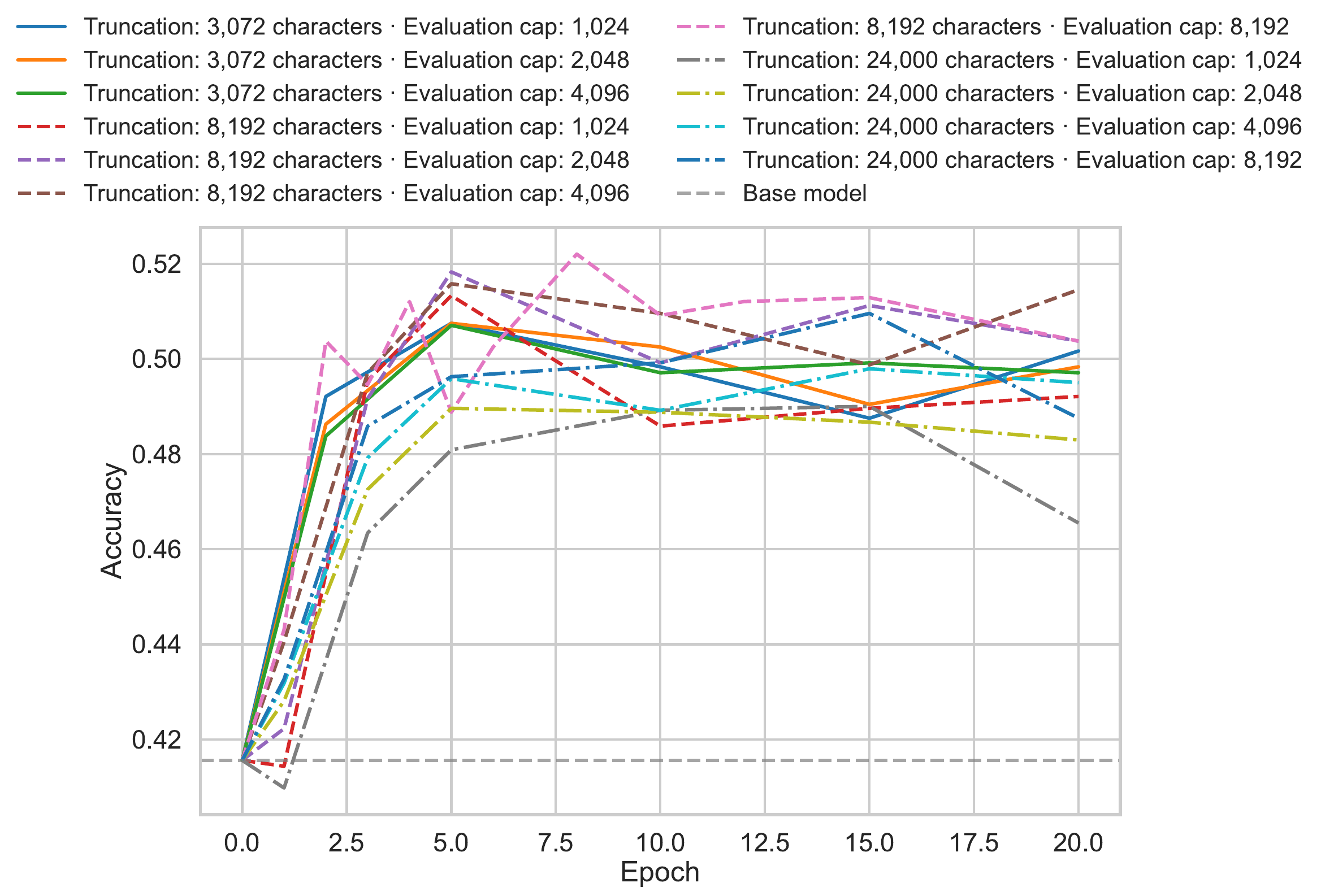}
    \caption{Head-truncation comparison}
  \end{subfigure}
  \hfill
  \begin{subfigure}[b]{0.32\linewidth}
    \centering
    \includegraphics[width=\linewidth]{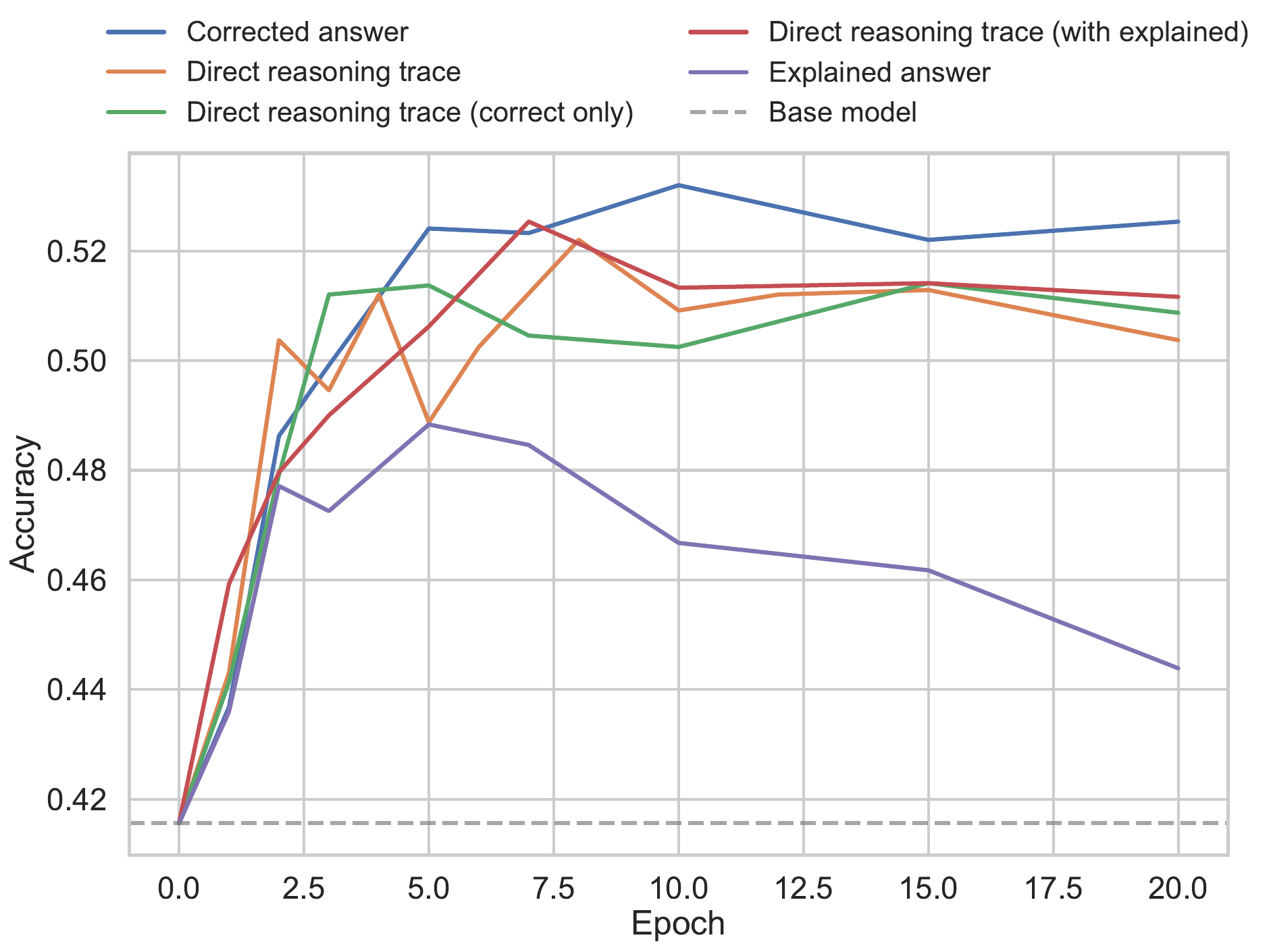}
    \caption{Generation methods: truncation 8,192, eval cap 8,192}
  \end{subfigure}
  \hfill
  \begin{subfigure}[b]{0.32\linewidth}
    \centering
    \includegraphics[width=\linewidth]{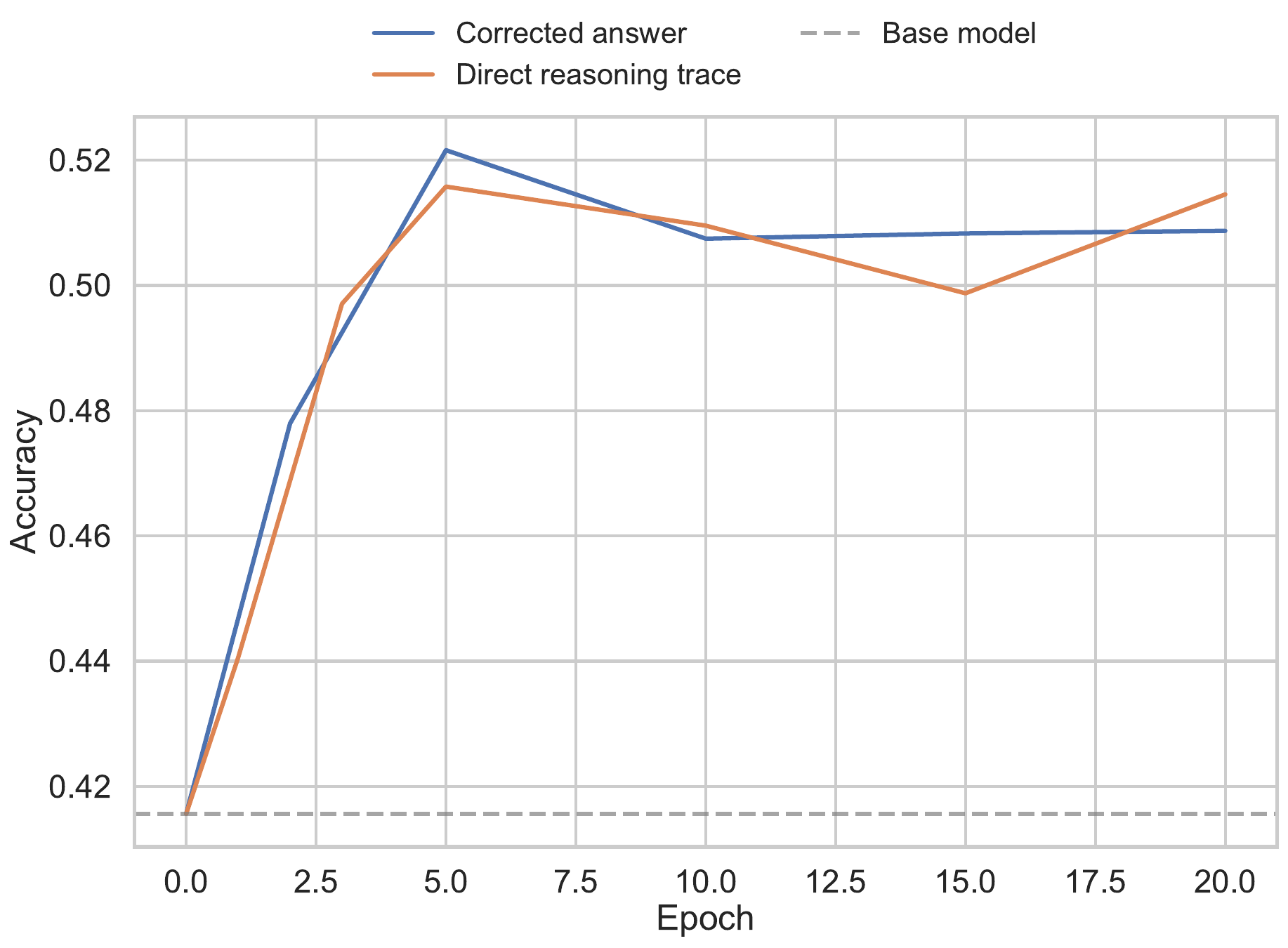}
    \caption{Generation methods: truncation 8,192, eval cap 4,096}
  \end{subfigure}
  \caption{Synthetic-trace distillation results for Qwen 2.5 3B.}
  \label{fig:synth_qwen_individual}
\end{figure*}

\begin{figure*}[t]
  \centering
  \begin{subfigure}[b]{0.48\linewidth}
    \centering
    \includegraphics[width=\linewidth]{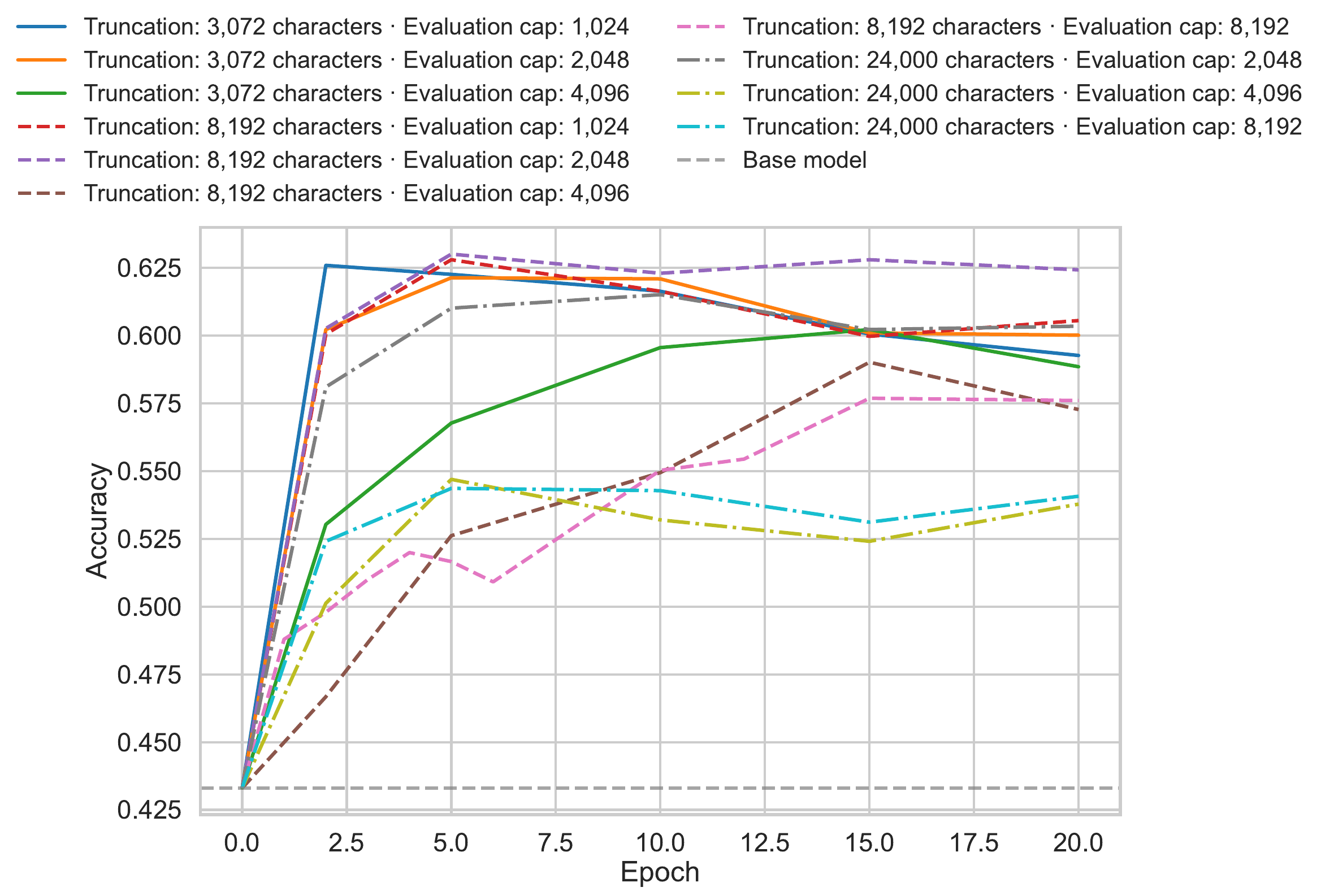}
    \caption{Head-truncation comparison}
  \end{subfigure}
  \hfill
  \begin{subfigure}[b]{0.48\linewidth}
    \centering
    \includegraphics[width=\linewidth]{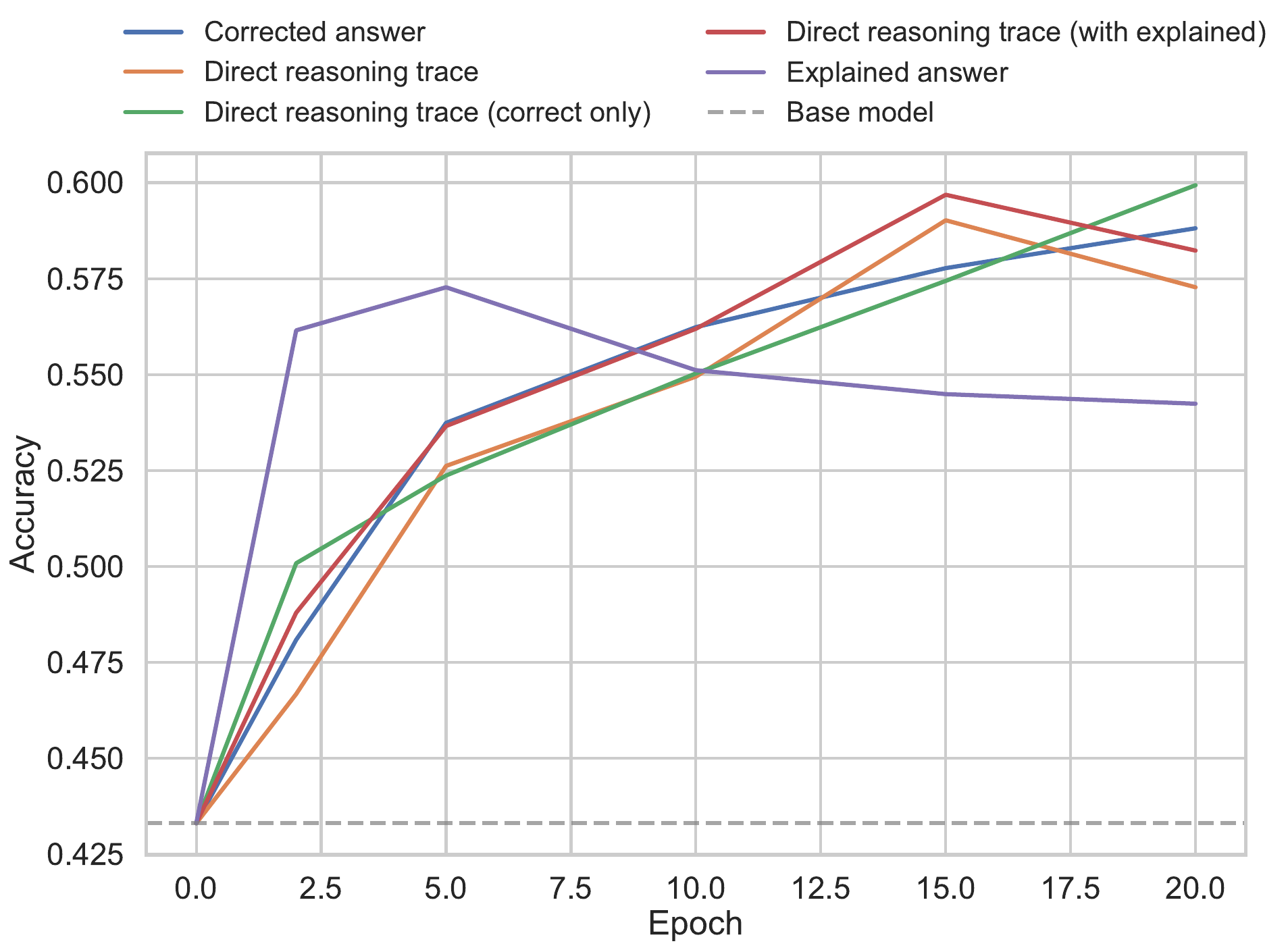}
    \caption{Generation methods: truncation 8,192, eval cap 4,096}
  \end{subfigure}
  \caption{Synthetic-trace distillation results for Phi4-mini.}
  \label{fig:synth_phi_individual}
\end{figure*}

Figures~\ref{fig:synth_llama_individual}--\ref{fig:synth_phi_individual} show test accuracy across training epochs for each student. The left panels compare head-truncation limits of 3,072, 8,192, and 24,000 characters for direct reasoning traces; each legend separately specifies the evaluation cap in tokens. The generation-method panels hold head truncation fixed at 8,192 characters and compare methods at the stated evaluation cap. Dashed horizontal lines and epoch-zero values use the base model's accuracy when prompted to answer with chain-of-thought (base models are not originally reasoners), shared across evaluation budgets.

\section{Synthetic trace truncation and reasoning caps}
\label{sec:trace_truncation_caps}

Training truncation limits the distilled teacher reasoning in characters. Reasoning cap limits the student's generated reasoning at evaluation time in tokens. They are evaluated independently.
At evaluation, reaching the reasoning cap forces a closing thinking token and allows generation to continue toward the answer, as detailed in Appendix~\ref{sec:mix_answer_formatting}.

Table~\ref{tab:training_trace_truncation_counts} reports how many training traces exceed each character budget before truncation.

\begin{table}[t]
  \centering
  \small
  \setlength{\tabcolsep}{4pt}
  \begin{tabular}{rrr}
    \toprule
    Budget (chars) & Direct traces & Corrected traces \\
    \midrule
    3,072 & 3,603 (37.4\%) & 3,721 (38.7\%) \\
    8,192 & 1,879 (19.5\%) & 1,921 (20.0\%) \\
    24,000 & 738 (7.7\%) & 749 (7.8\%) \\
    \bottomrule
  \end{tabular}
  \caption{Number (percentage) of training traces exceeding each truncation budget}
  \label{tab:training_trace_truncation_counts}
\end{table}

\subsection{Which part of the trace is retained?}
Head truncation keeps a prefix. Middle truncation keeps an initial segment and a final segment. Tail truncation keeps only a suffix. At 24,000-character limit, head retains up to 24,000 characters, middle retains the first 18,000 and last 6,000 of overlong traces, and tail retains the last 24,000. For corrected traces, middle truncation instead reserves up to 1,600 characters for the correction and uses the remaining space for the original prefix.

\begin{figure*}[t]
  \centering
  \begin{subfigure}[b]{0.32\linewidth}
    \centering
    \includegraphics[width=\linewidth]{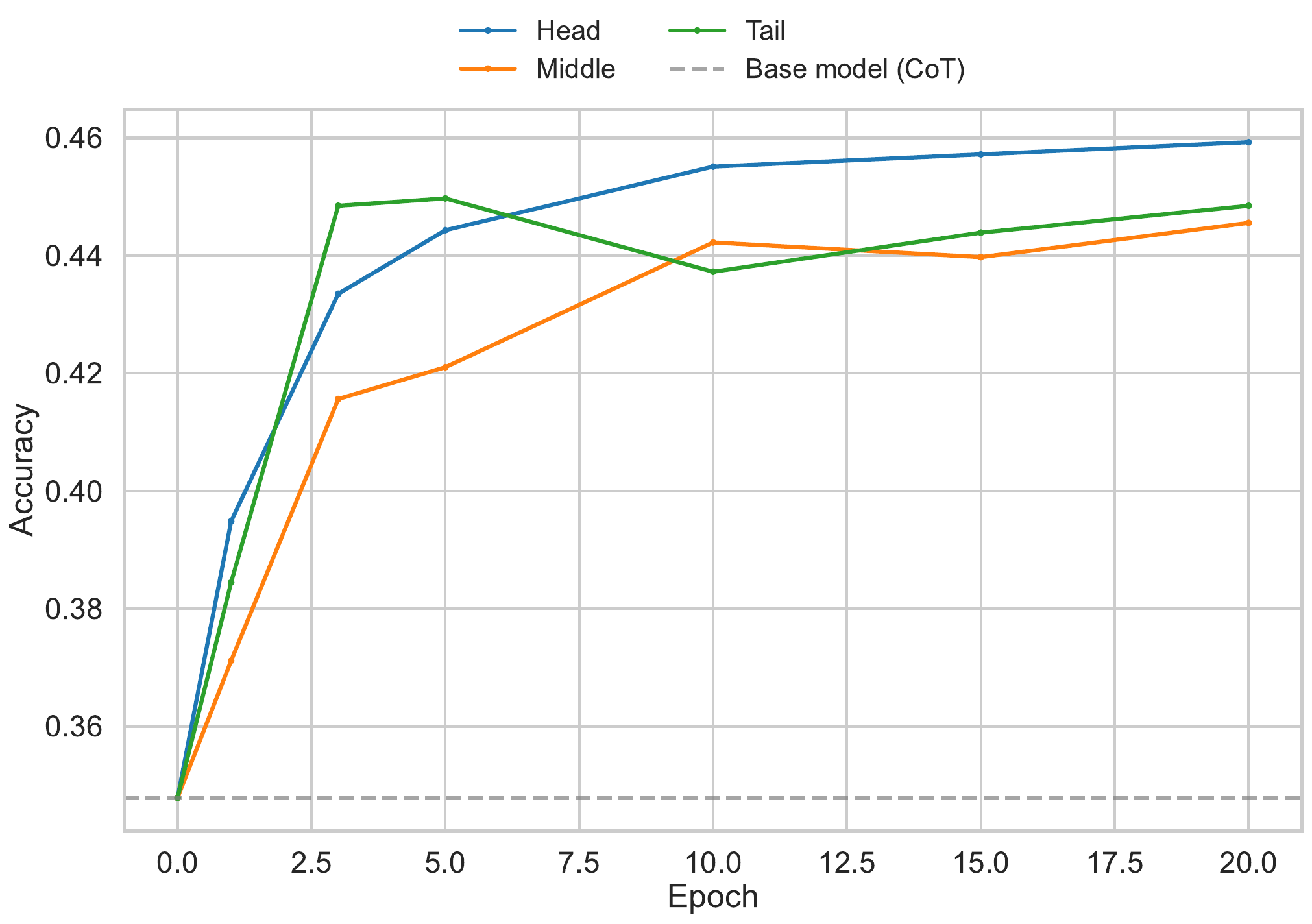}
    \caption{LLaMA 3.2 3B}
  \end{subfigure}
  \hfill
  \begin{subfigure}[b]{0.32\linewidth}
    \centering
    \includegraphics[width=\linewidth]{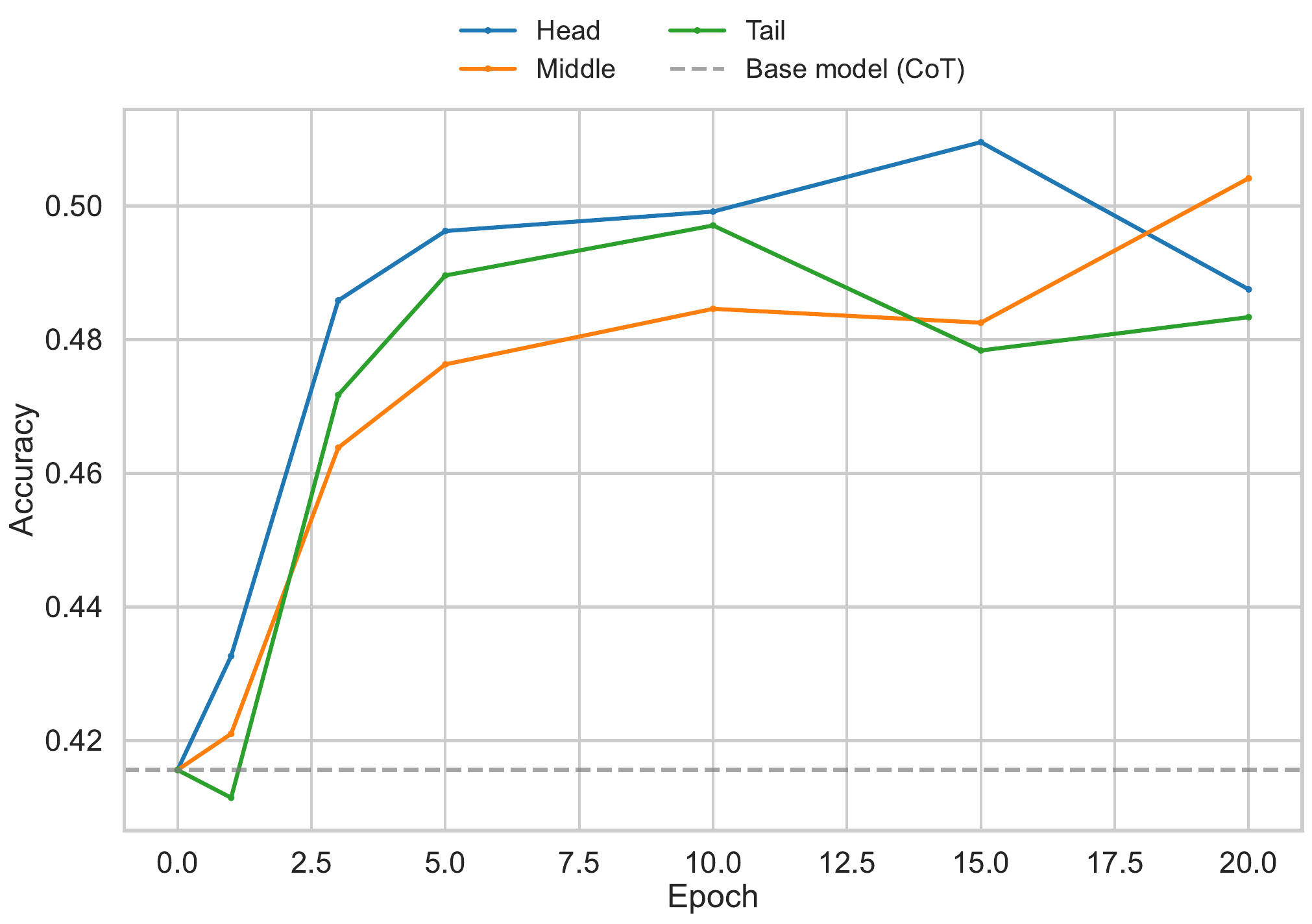}
    \caption{Qwen 2.5 3B}
  \end{subfigure}
  \hfill
  \begin{subfigure}[b]{0.32\linewidth}
    \centering
    \includegraphics[width=\linewidth]{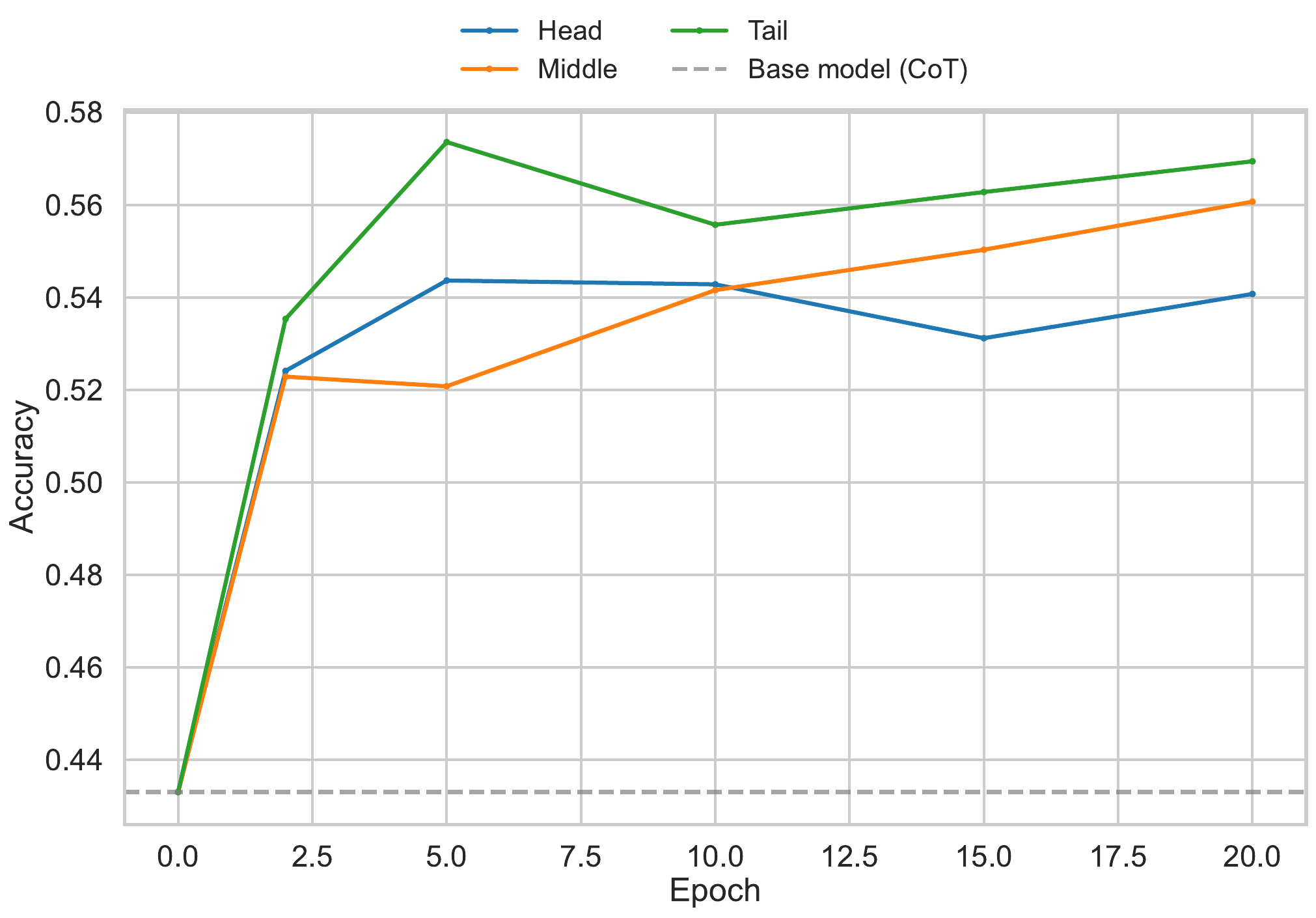}
    \caption{Phi4-mini}
  \end{subfigure}
  \caption{Training truncation policies at the nominal 24,000-character setting, evaluated with an 8,192-token reasoning cap. Dashed lines and epoch-zero points give the base model's chain-of-thought accuracy.}
  \label{fig:trace_truncation_modes}
\end{figure*}

Figure~\ref{fig:trace_truncation_modes} compares these policies at a fixed 8,192-token evaluation cap.

\subsection{Training truncation length versus evaluation cap}
\begin{figure*}[t]
  \centering
  \begin{subfigure}[b]{0.32\linewidth}
    \centering
    \includegraphics[width=\linewidth]{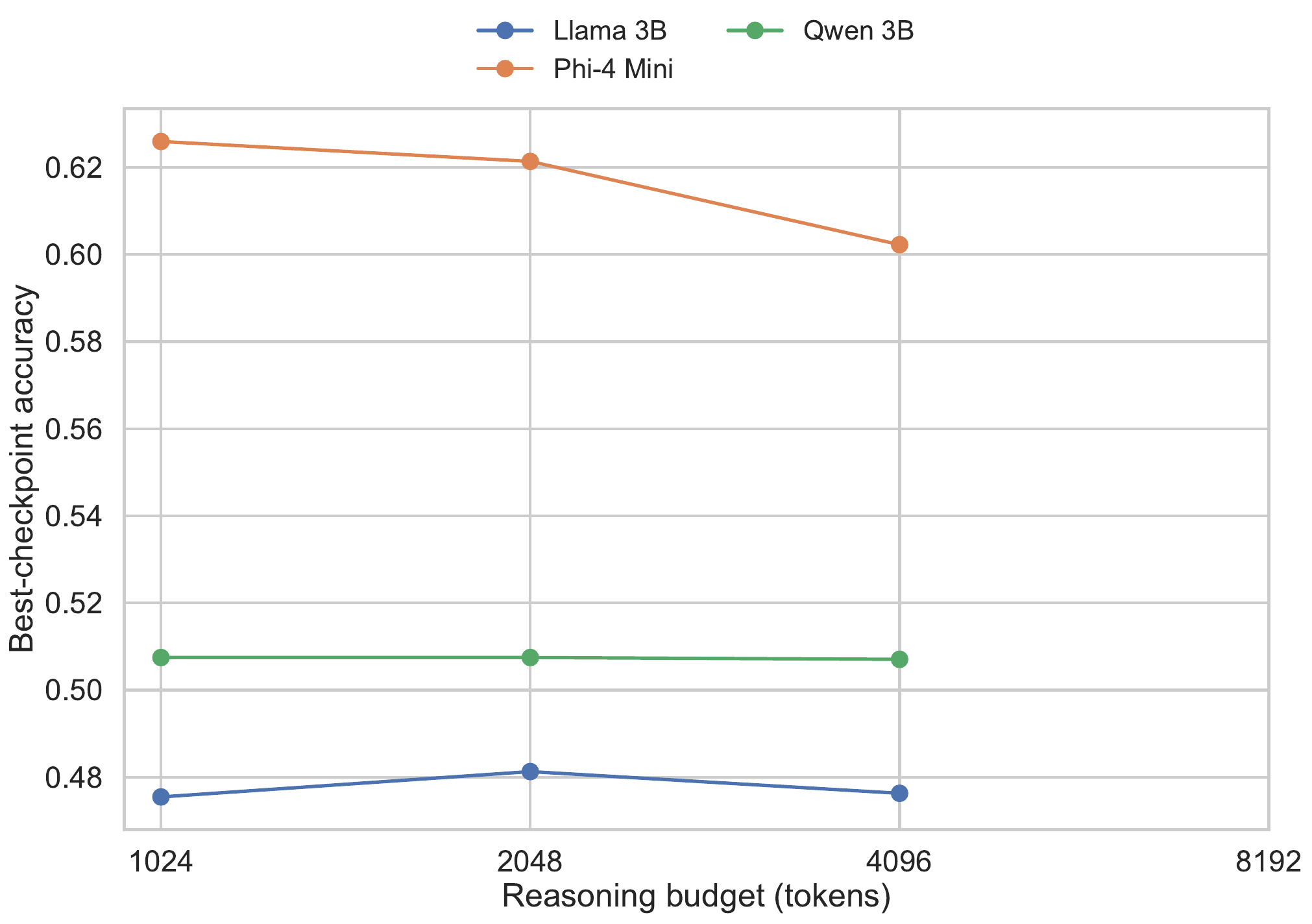}
    \caption{Head truncation: 3,072 characters}
  \end{subfigure}
  \hfill
  \begin{subfigure}[b]{0.32\linewidth}
    \centering
    \includegraphics[width=\linewidth]{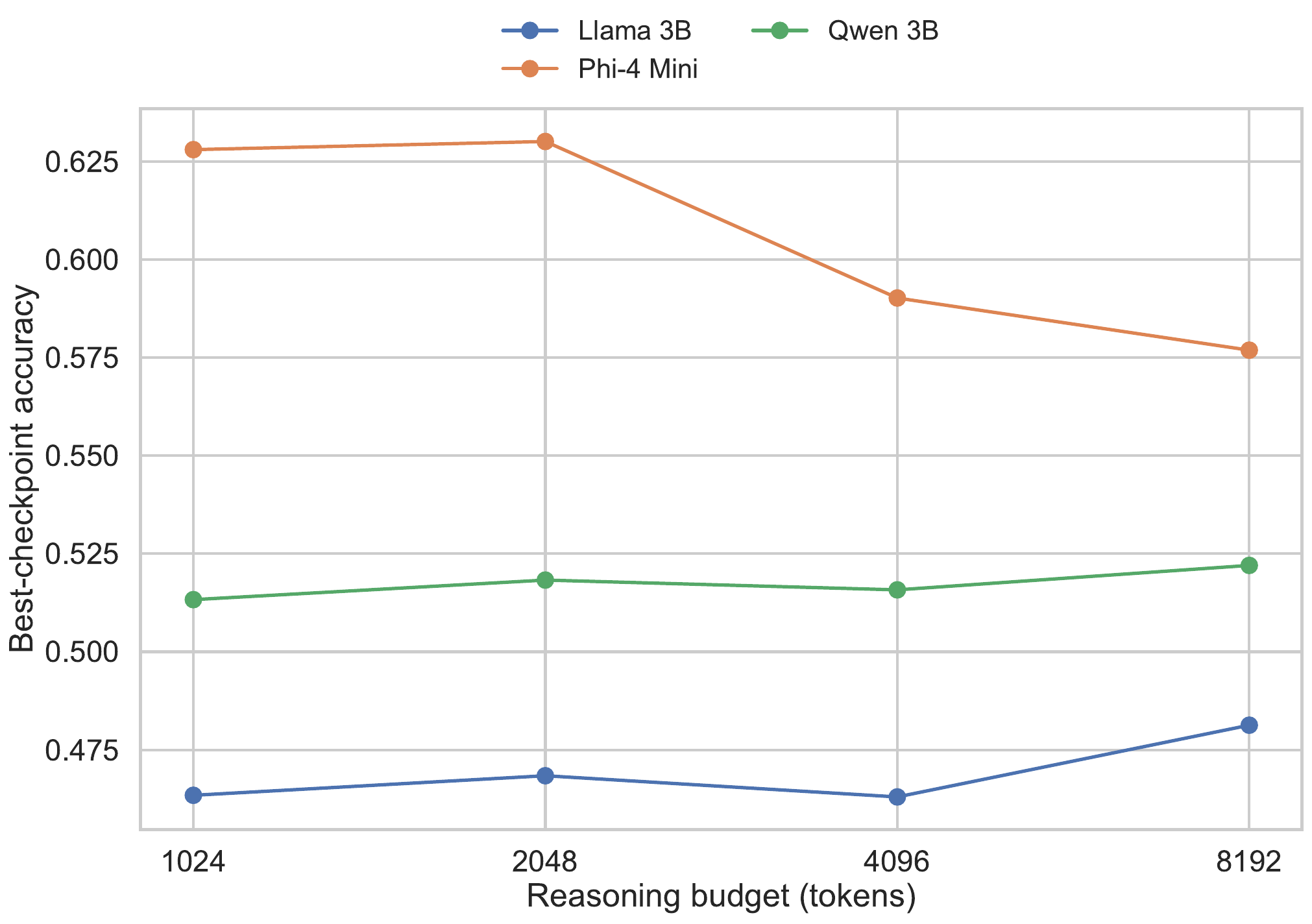}
    \caption{Head truncation: 8,192 characters}
  \end{subfigure}
  \hfill
  \begin{subfigure}[b]{0.32\linewidth}
    \centering
    \includegraphics[width=\linewidth]{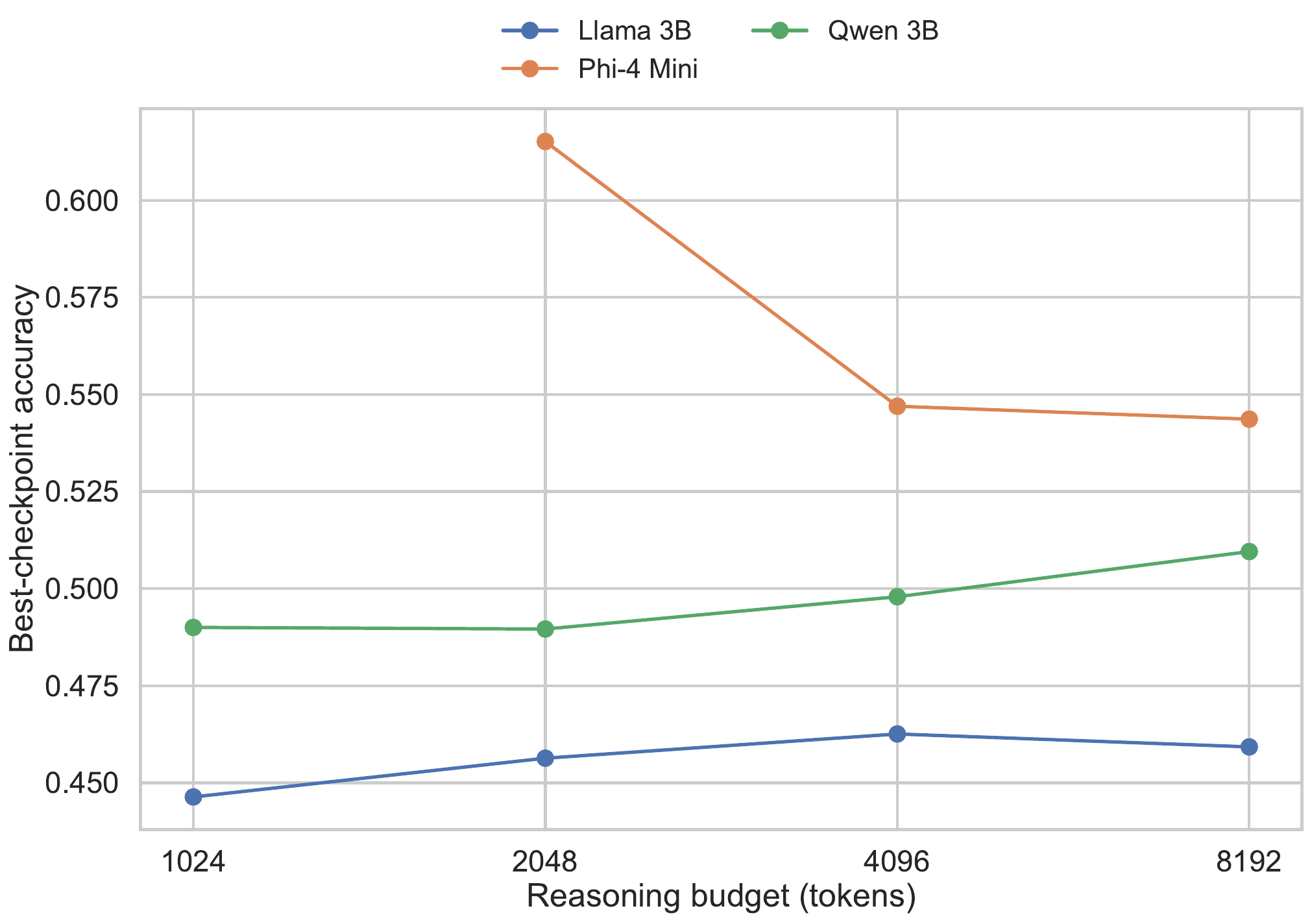}
    \caption{Head truncation: 24,000 characters}
  \end{subfigure}
  \caption{Best-checkpoint accuracy versus evaluation reasoning cap for each training truncation length.}
  \label{fig:trace_caps_best}
\end{figure*}

\begin{figure*}[t]
  \centering
  \begin{subfigure}[b]{0.32\linewidth}
    \centering
    \includegraphics[width=\linewidth]{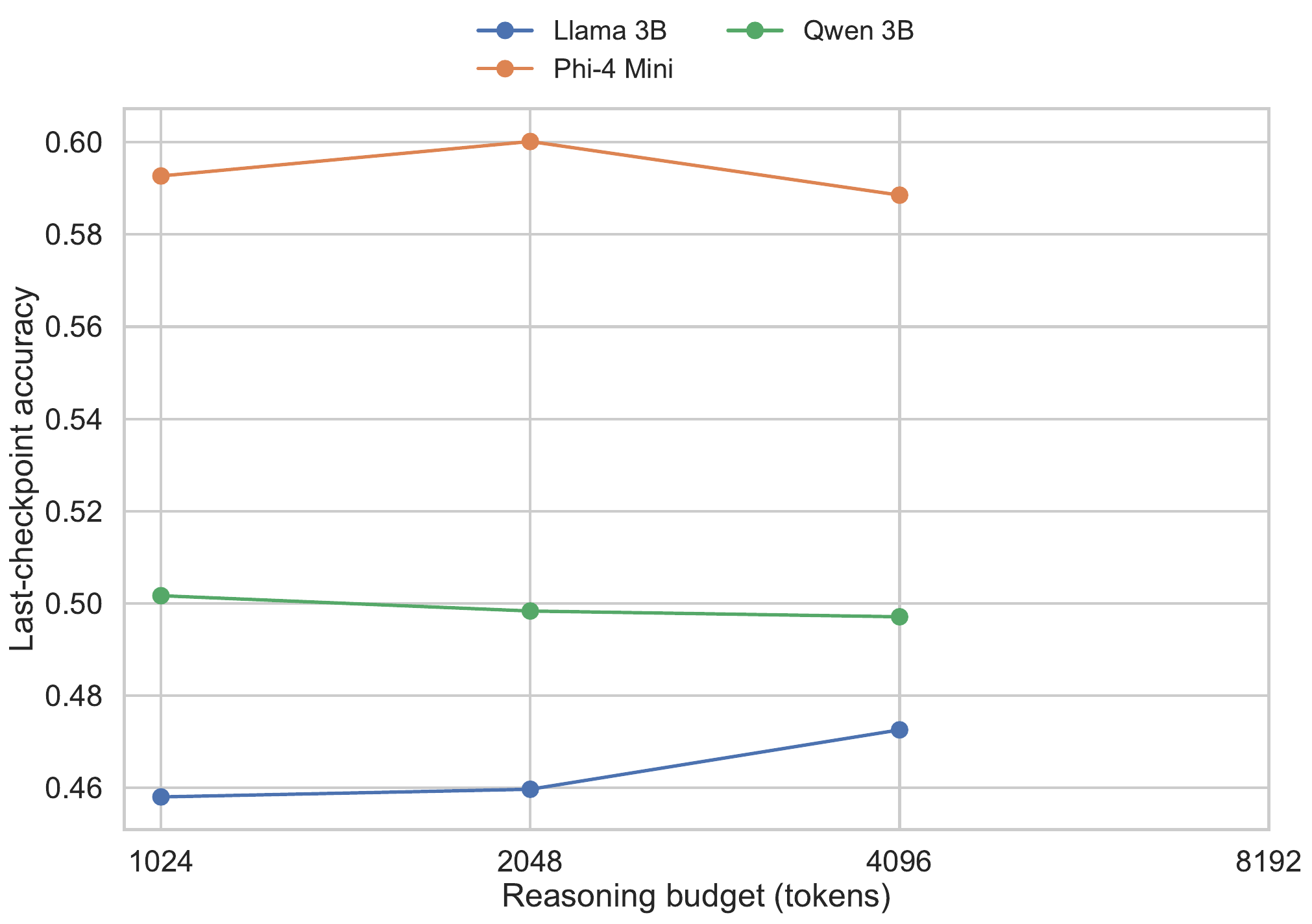}
    \caption{Head truncation: 3,072 characters}
  \end{subfigure}
  \hfill
  \begin{subfigure}[b]{0.32\linewidth}
    \centering
    \includegraphics[width=\linewidth]{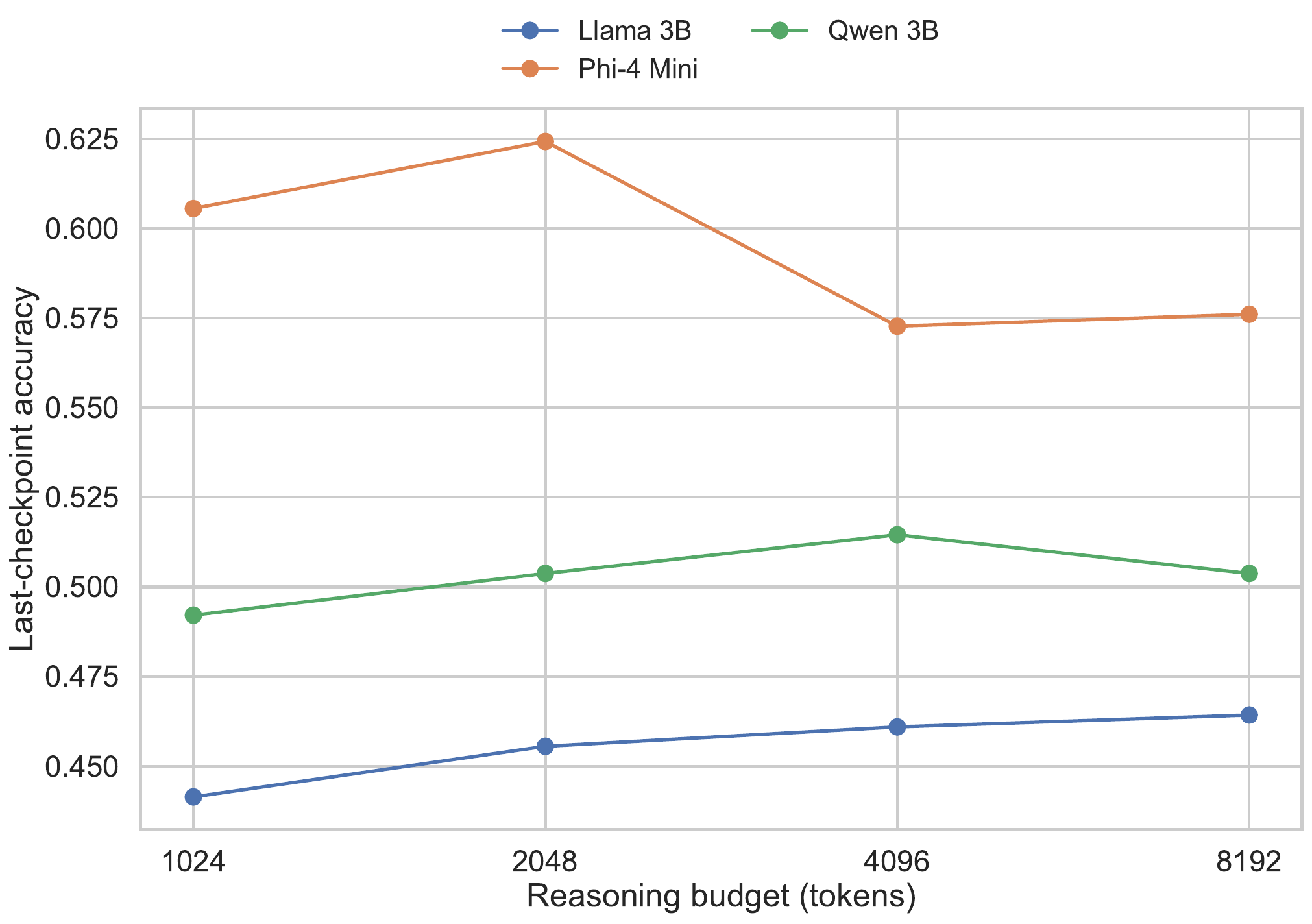}
    \caption{Head truncation: 8,192 characters}
  \end{subfigure}
  \hfill
  \begin{subfigure}[b]{0.32\linewidth}
    \centering
    \includegraphics[width=\linewidth]{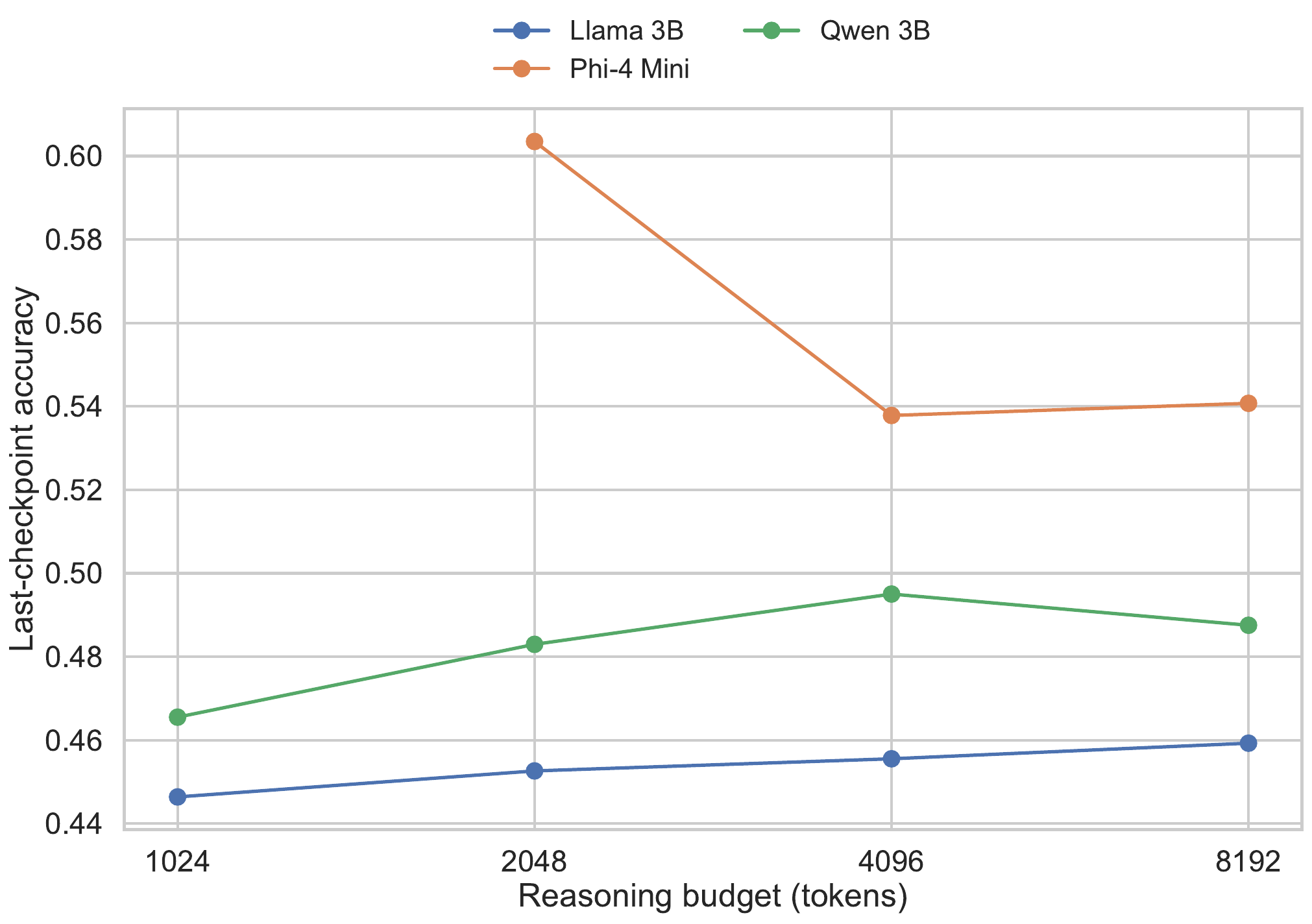}
    \caption{Head truncation: 24,000 characters}
  \end{subfigure}
  \caption{Last-checkpoint accuracy versus evaluation reasoning cap for each training truncation length.}
  \label{fig:trace_caps_last}
\end{figure*}

For direct traces with head truncation at 3,072, 8,192, or 24,000 characters, we compare available evaluation caps of 1,024, 2,048, 4,096, and 8,192 tokens. Figures~\ref{fig:trace_caps_best} and \ref{fig:trace_caps_last} separate training length from evaluation budget.

\section{Sensitivity Study Detailed Results}
\label{sec:sensitivity_details}

\begin{figure*}[t]
  \centering
  \begin{subfigure}[b]{0.32\linewidth}
    \centering
    \includegraphics[width=\linewidth]{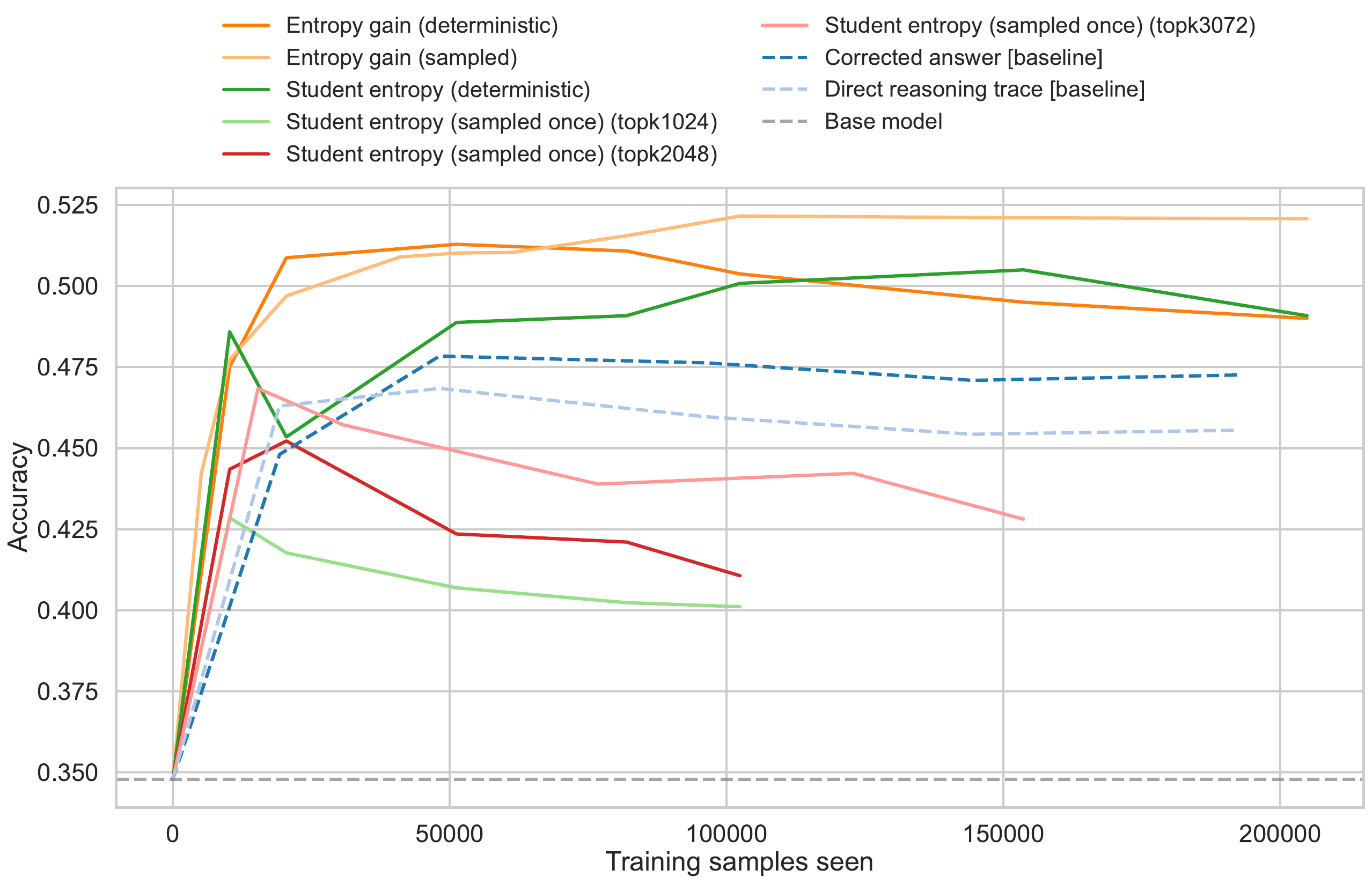}
    \caption{2,048-token thinking budget}
  \end{subfigure}
  \hfill
  \begin{subfigure}[b]{0.32\linewidth}
    \centering
    \includegraphics[width=\linewidth]{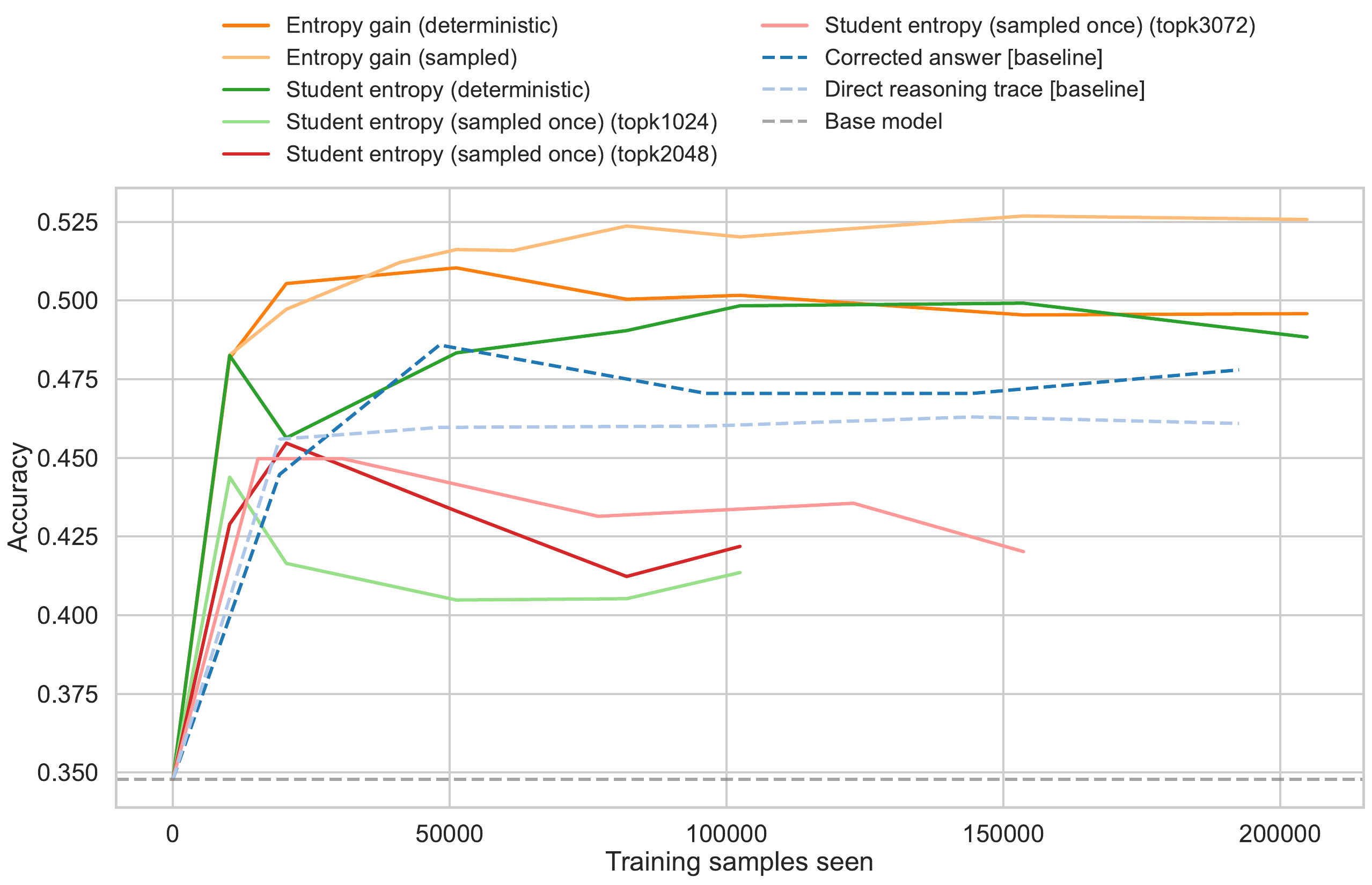}
    \caption{4,096-token thinking budget}
  \end{subfigure}
  \hfill
  \begin{subfigure}[b]{0.32\linewidth}
    \centering
    \includegraphics[width=\linewidth]{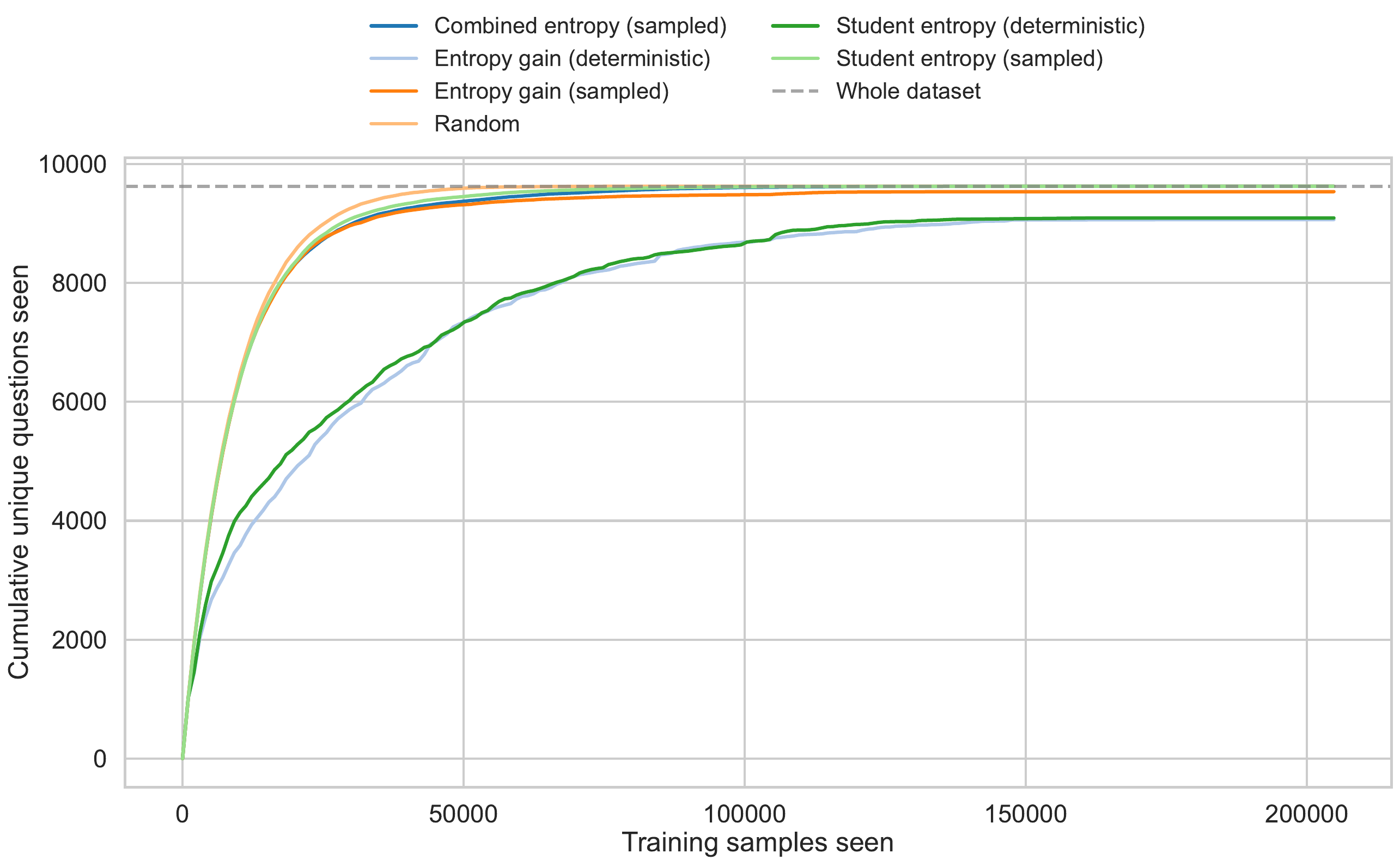}
    \caption{Cumulative unique question coverage}
  \end{subfigure}
  \caption{Distillation by sampling strategy for LLaMA 3.2 3B.}
  \label{fig:sensitivity_llama}
\end{figure*}

\begin{figure*}[t]
  \centering
  \begin{subfigure}[b]{0.32\linewidth}
    \centering
    \includegraphics[width=\linewidth]{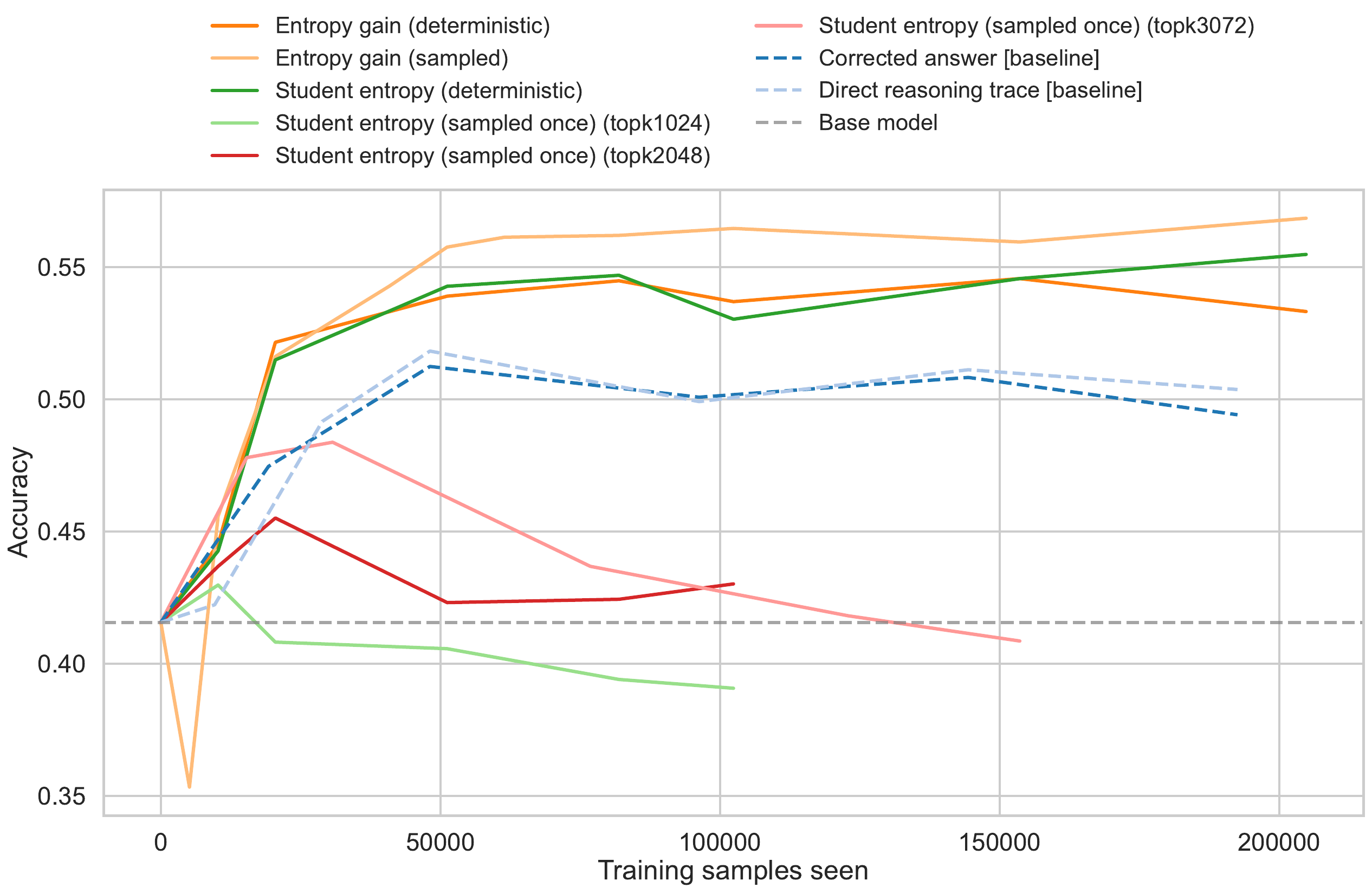}
    \caption{2,048-token thinking budget}
  \end{subfigure}
  \hfill
  \begin{subfigure}[b]{0.32\linewidth}
    \centering
    \includegraphics[width=\linewidth]{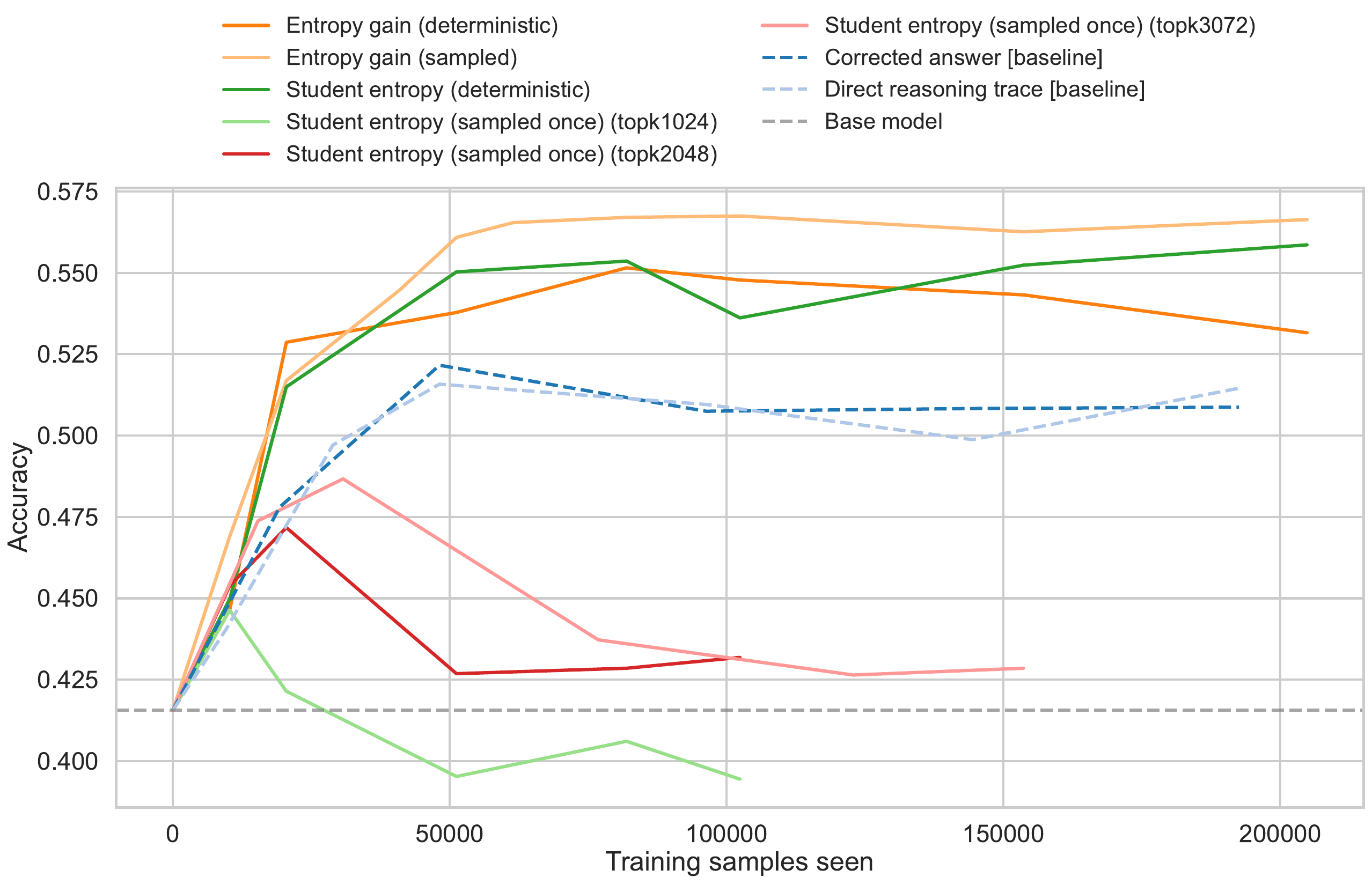}
    \caption{4,096-token thinking budget}
  \end{subfigure}
  \hfill
  \begin{subfigure}[b]{0.32\linewidth}
    \centering
    \includegraphics[width=\linewidth]{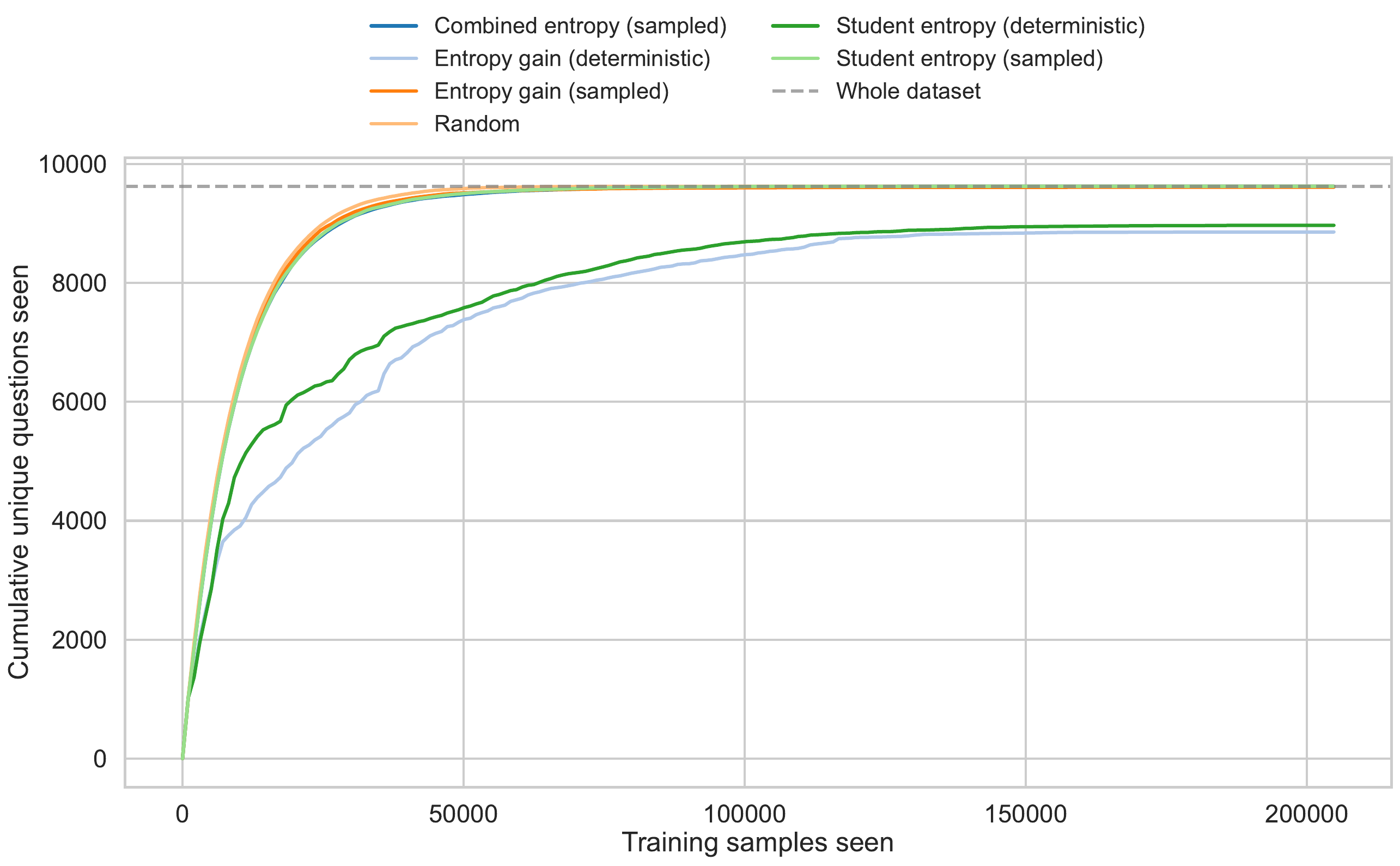}
    \caption{Cumulative unique question coverage}
  \end{subfigure}
  \caption{Distillation by sampling strategy for Qwen 2.5 3B.}
  \label{fig:sensitivity_qwen}
\end{figure*}

\begin{figure*}[t]
  \centering
  \begin{subfigure}[b]{0.32\linewidth}
    \centering
    \includegraphics[width=\linewidth]{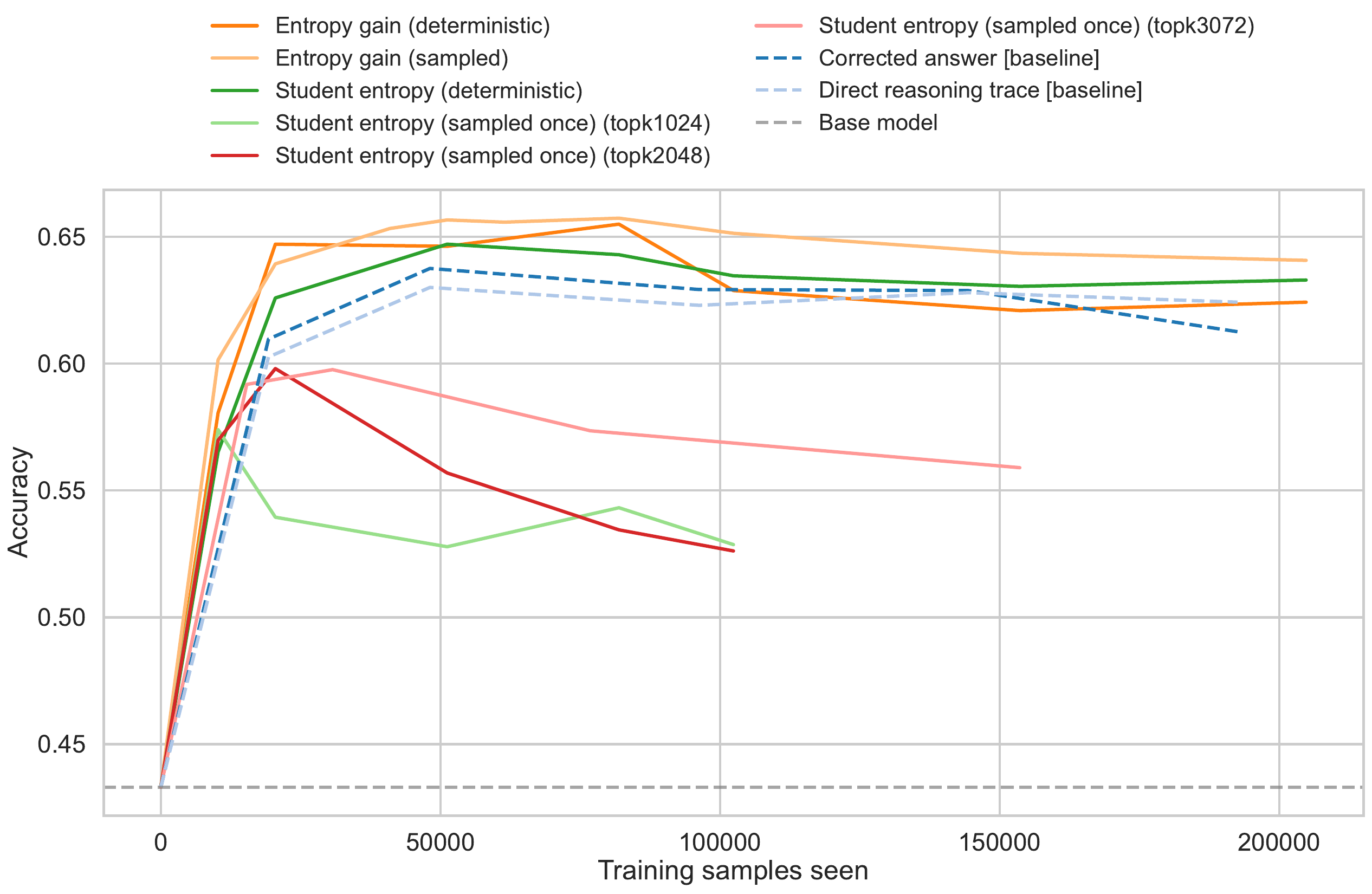}
    \caption{2,048-token thinking budget}
  \end{subfigure}
  \hfill
  \begin{subfigure}[b]{0.32\linewidth}
    \centering
    \includegraphics[width=\linewidth]{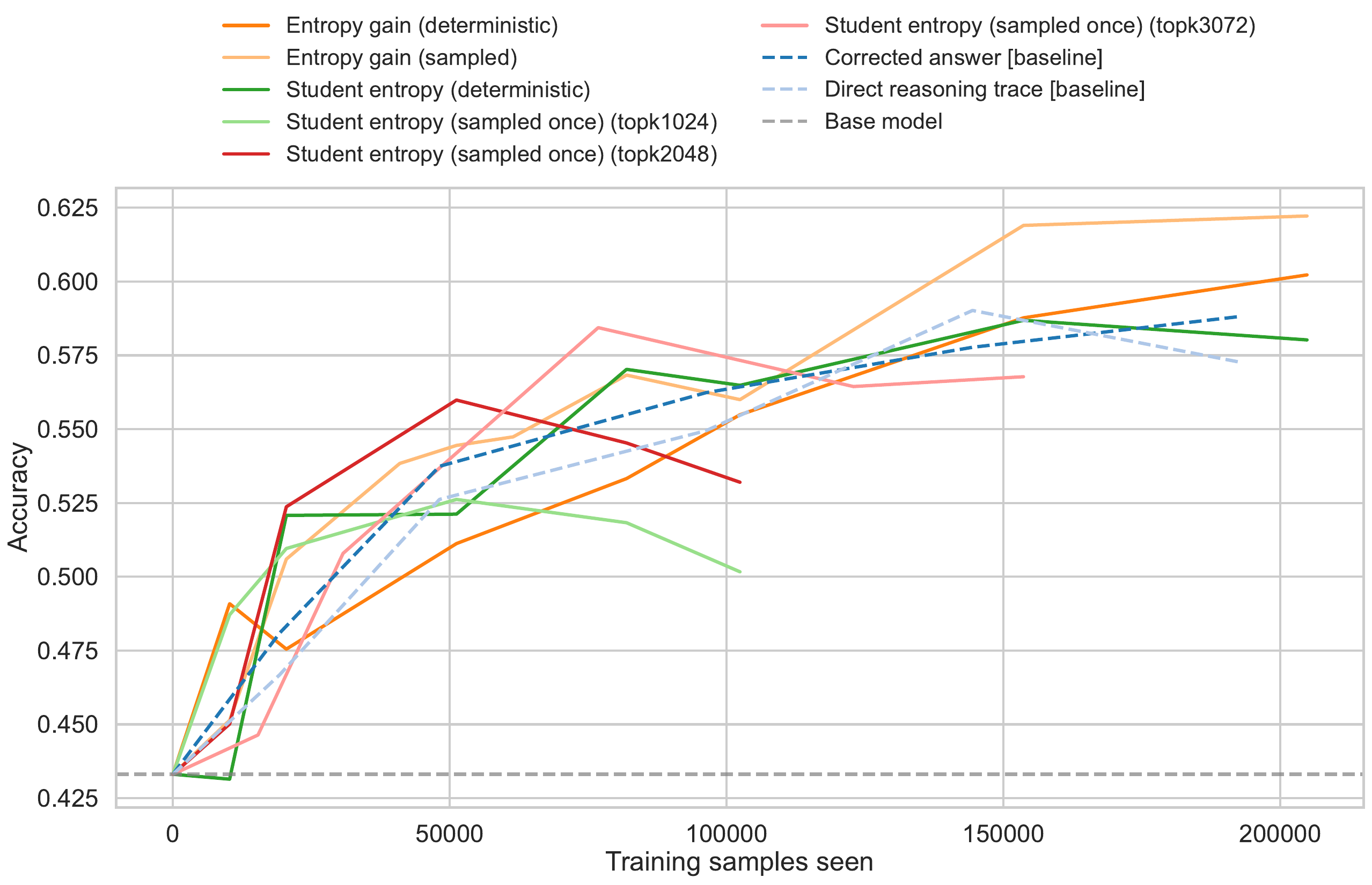}
    \caption{4,096-token thinking budget}
  \end{subfigure}
  \hfill
  \begin{subfigure}[b]{0.32\linewidth}
    \centering
    \includegraphics[width=\linewidth]{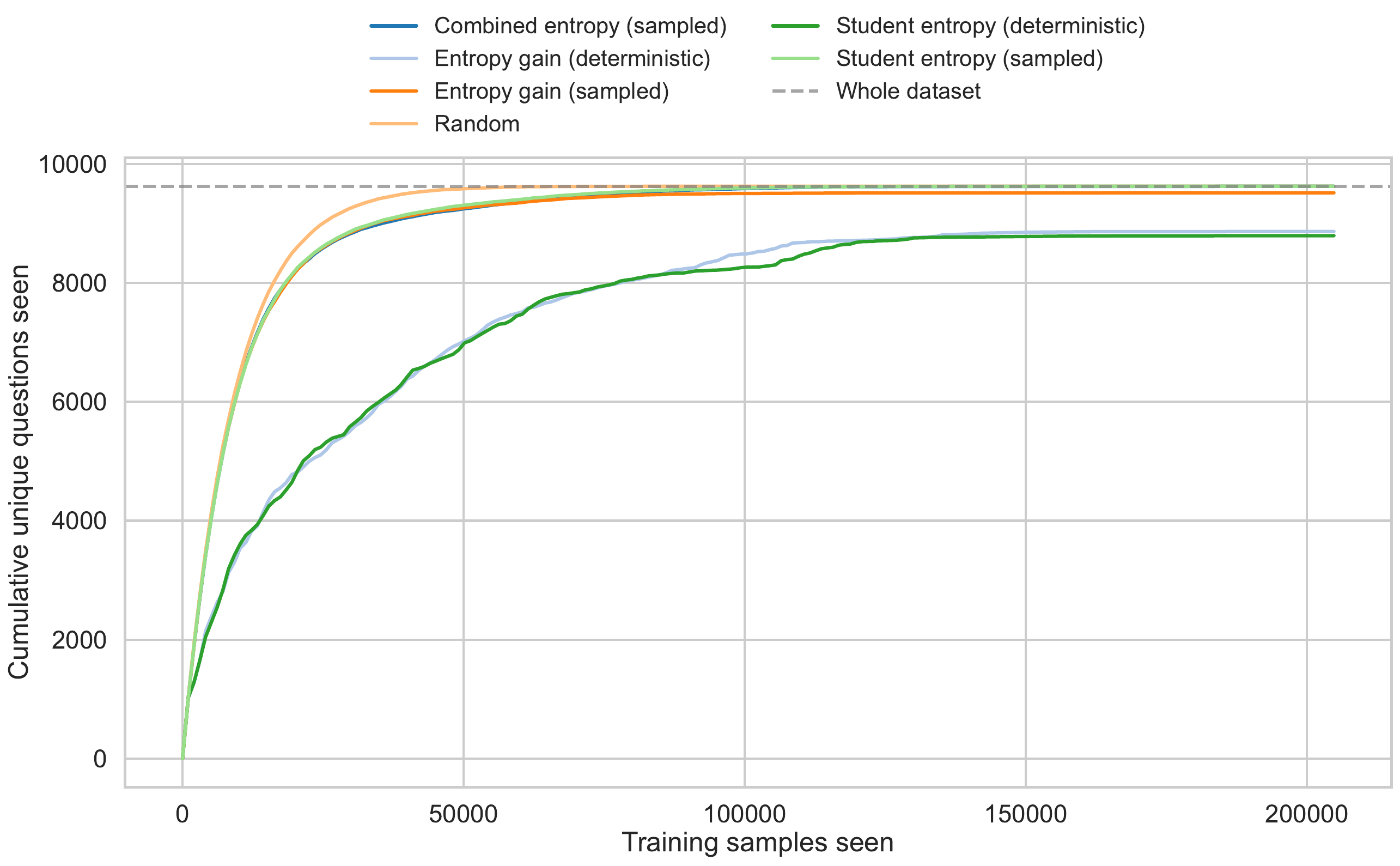}
    \caption{Cumulative unique question coverage}
  \end{subfigure}
  \caption{Distillation by sampling strategy for Phi4-mini.}
  \label{fig:sensitivity_phi}
\end{figure*}

\begin{figure*}[t]
  \centering
  \begin{subfigure}[b]{0.32\linewidth}
    \centering
    \includegraphics[width=\linewidth]{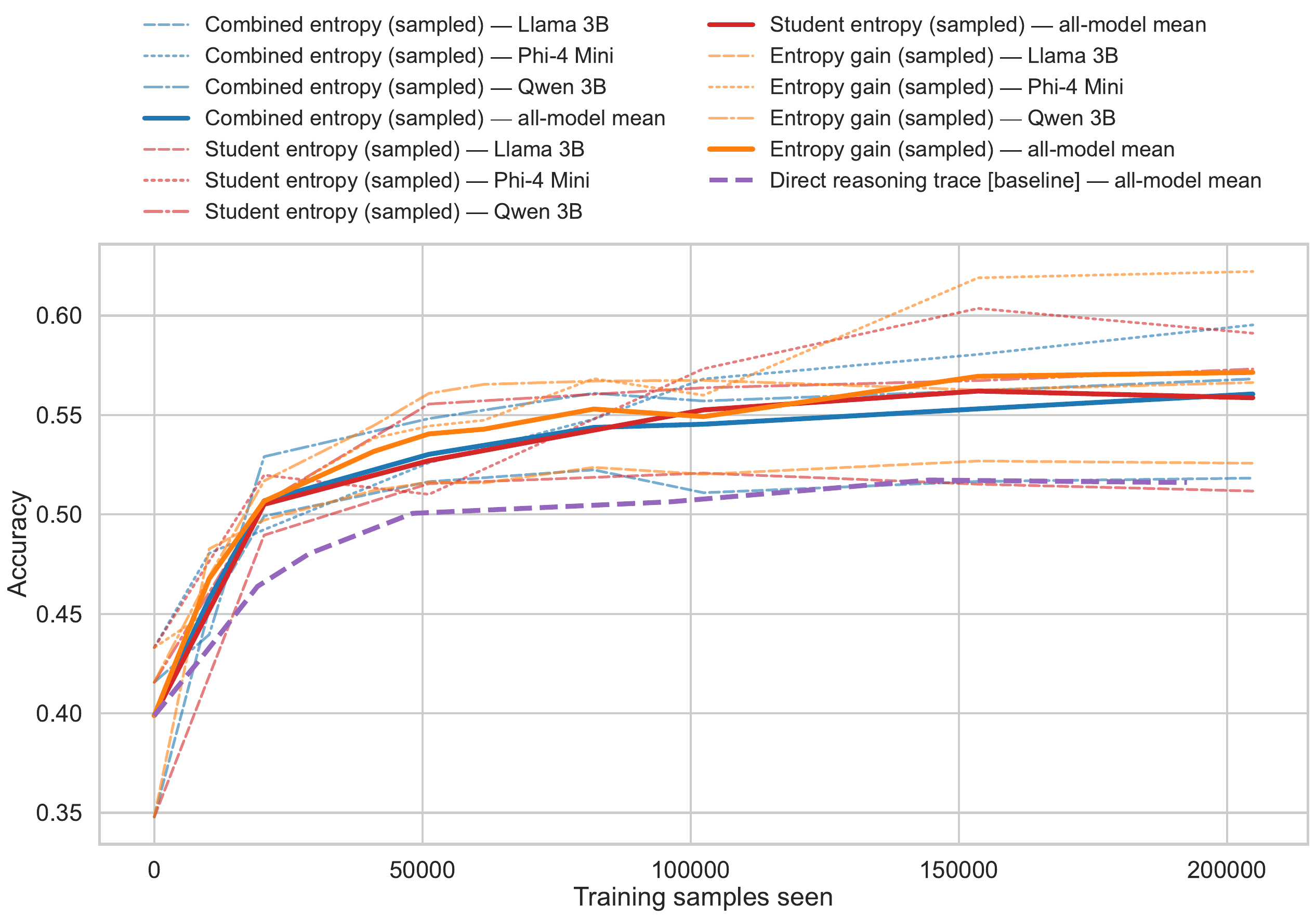}
    \caption{Weighted samplers, 4,096-token thinking budget}
  \end{subfigure}
  \hfill
  \begin{subfigure}[b]{0.32\linewidth}
    \centering
    \includegraphics[width=\linewidth]{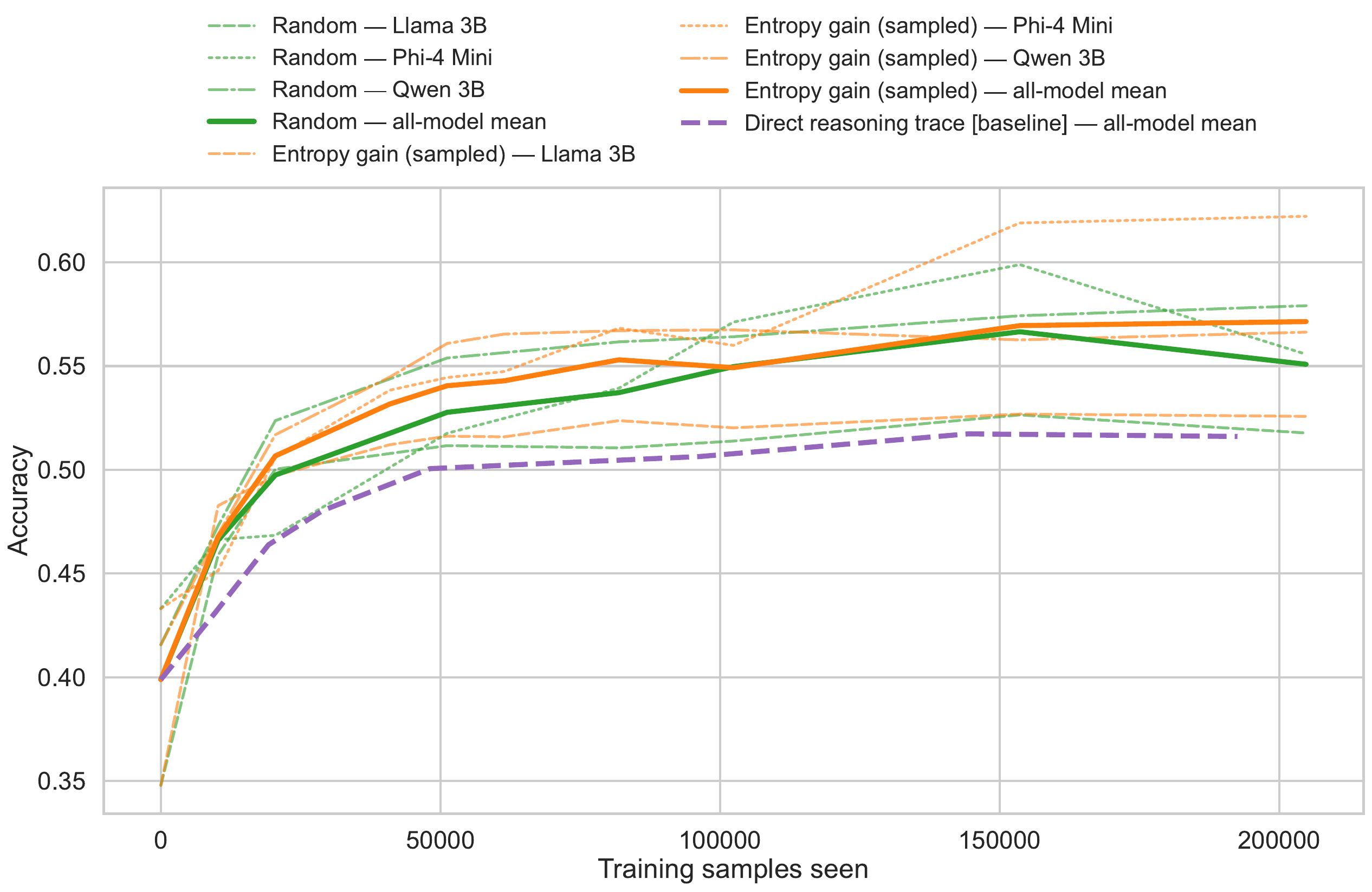}
    \caption{Random versus entropy gain, 4,096-token thinking budget}
  \end{subfigure}
  \hfill
  \begin{subfigure}[b]{0.32\linewidth}
    \centering
    \includegraphics[width=\linewidth]{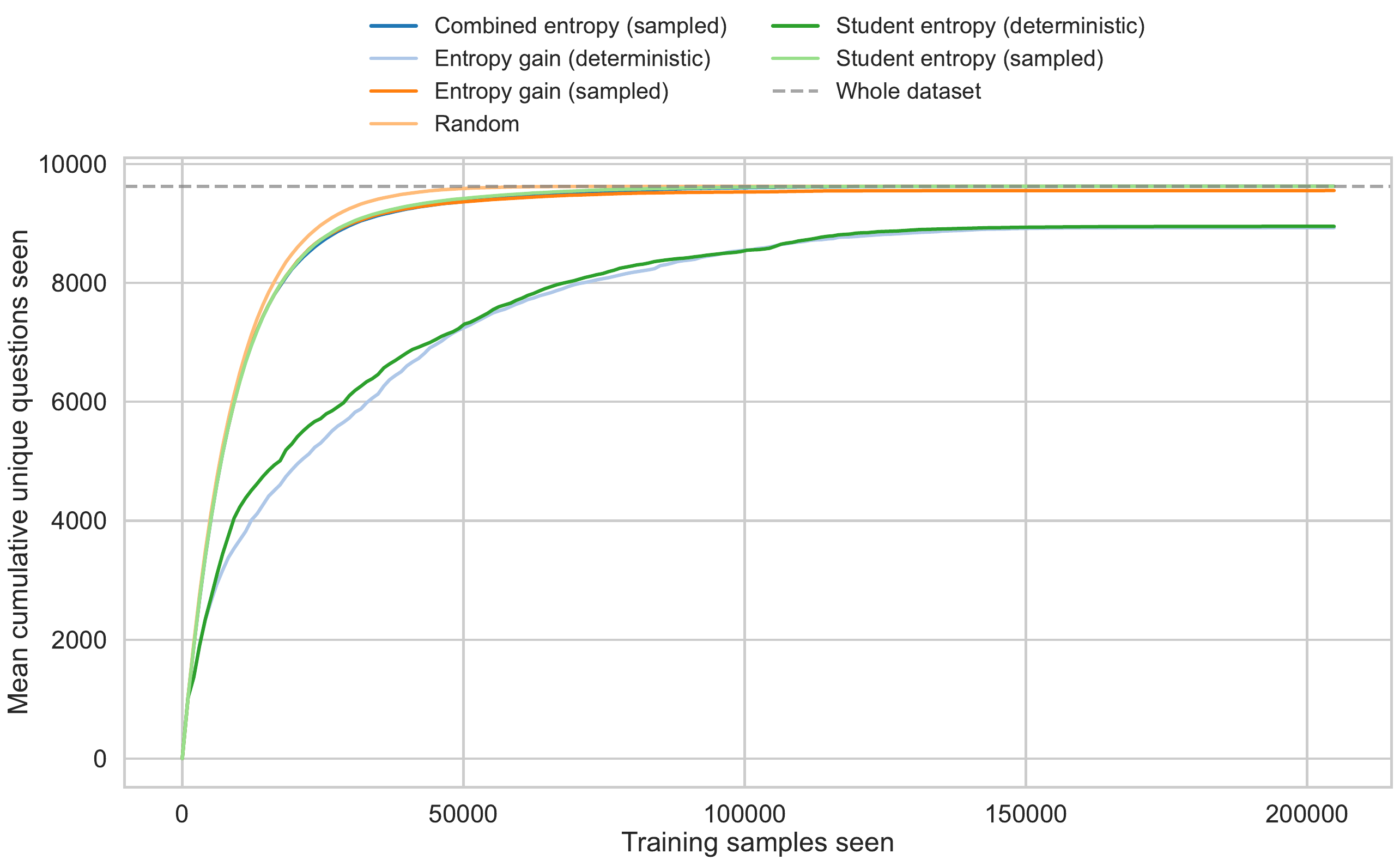}
    \caption{Mean unique question coverage}
  \end{subfigure}
  \caption{Sampler comparisons and question coverage across all three models. In the accuracy panels, thin dashed curves show individual models, thick solid curves their equal-weight means, and the thick dashed curve the direct-trace baseline mean.}
  \label{fig:sensitivity_sampler_comparisons}
\end{figure*}

Figure~\ref{fig:sensitivity_sampler_comparisons} compares the weighted samplers and random sampling.
Figures~\ref{fig:sensitivity_llama}--\ref{fig:sensitivity_phi} provide individual accuracy and coverage curves.
Coverage counts cumulative distinct question IDs selected at least once, so repeated occurrences do not increase it.

\clearpage
\raggedbottom
\section{Cross-model diagnostics of the uncertainty proxy}
\label{sec:uncertainty_proxy_diagnostics}

We compare single-token entropies and accuracies of both uncertainty proxies and all three base students before distillation.
Accuracy measures single-token answers rather than the reasoning accuracy reported in the main results.
Both proxies are substantially more accurate than the students (Table~\ref{tab:uncertainty_proxy_models}).

\begin{table}[H]
  \centering
  \small
  \setlength{\tabcolsep}{4pt}
  \begin{tabular}{lrr}
    \toprule
    Model & Accuracy (\%) & Mean $H$ \\
    \midrule
    LLaMA 70B & 52.2 & 0.0126 \\
    Qwen 72B & 54.2 & 0.0265 \\
    LLaMA 3B & 28.9 & 0.1189 \\
    Qwen 3B & 36.2 & 0.0287 \\
    Phi4-mini & 37.8 & 0.1101 \\
    \bottomrule
  \end{tabular}
  \caption{Single-token accuracy and mean normalized entropy on the full MMLU-Pro dataset}
  \label{tab:uncertainty_proxy_models}
\end{table}

We also analyze agreement and correlation between entropies. Table~\ref{tab:uncertainty_proxy_correlations} compares entropy values and correctness using Pearson $r$, and entropy rankings using Spearman $\rho$.

\begin{table}[H]
  \centering
  \small
  \setlength{\tabcolsep}{2pt}
  \begin{tabular}{lrrr}
    \toprule
    & \multicolumn{2}{c}{Entropy} & Correctness \\
    \cmidrule(lr){2-3}
    Models compared & $r$ & $\rho$ & $r$ \\
    \midrule
    \textbf{LLaMA 70B / Qwen 72B} & \textbf{0.240} & \textbf{0.533} & \textbf{0.554} \\
    \midrule
    Mean proxy / LLaMA 3B & 0.369 & 0.435 & 0.322 \\
    Mean proxy / Qwen 3B & 0.343 & 0.461 & 0.447 \\
    Mean proxy / Phi4-mini & 0.502 & 0.582 & 0.441 \\
    \bottomrule
  \end{tabular}
  \caption{Pearson $r$ for entropy values and correctness, and Spearman $\rho$ for entropy rankings. The first row compares the proxies; remaining rows compare their mean entropy and mean correctness scores with each student's scores on the common test split.}
  \label{tab:uncertainty_proxy_correlations}
\end{table}

\paragraph{Interpretation.}
Table~\ref{tab:uncertainty_proxy_correlations} shows moderate agreement in proxy entropy rankings, with weaker linear agreement, which suggests aggregating entropies across models.
Correctness agreement is higher between proxies than between the mean proxy and any student. It suggests a greater shared knowledge signal.
The largest correctness gap occurs for LLaMA 3B, which also shows the largest uplift over direct distillation (Table~\ref{tab:distillation_main_results}); Phi4-mini's closer entropy rankings could help explain its smaller uplift.
These patterns are consistent with a shared difficulty baseline alongside student-specific uncertainty.
Under the assumption in Section~\ref{sec:method} that larger models have lower total uncertainty, they motivate using larger-model entropy to estimate the knowledge gap.

\end{document}